\documentclass[11pt]{article}

\usepackage[preprint]{acl}

\usepackage{times}
\usepackage{latexsym}
\usepackage{amsmath,amssymb}
\usepackage{array}
\usepackage{tabularx}
\usepackage{booktabs}
\usepackage{multirow}
\usepackage{float}
\usepackage{dblfloatfix}
\usepackage{needspace}
\usepackage[T1]{fontenc}

\usepackage[utf8]{inputenc}

\usepackage{microtype}

\usepackage{inconsolata}

\usepackage{graphicx}

\newcommand{\mergedgain}[2]{\makebox[0pt][r]{\raisebox{-0.45ex}{\textcolor{red}{\tiny(+#2)}}\,}\textbf{#1}}

\newcommand{\bestcorner}[2]{\makebox[0pt][r]{\raisebox{-0.45ex}{\textcolor{red}{\tiny(+#2)}}\,}\textbf{#1}}

\title{Rethinking Heterogeneous LLM Merging: A Weighted Model Averaging Perspective}

\author{
Jiahe Fan\textsuperscript{1} \quad
Yinghao Hou\textsuperscript{1} \quad
Si Chen\textsuperscript{2} \quad
Aiyuan Zhang\textsuperscript{1} \\
Hong Xie\textsuperscript{3} \quad
Defu Lian\textsuperscript{3} \\
\textsuperscript{1}University of Science and Technology of China \\
\textsuperscript{2}School of Information Science and Technology, Department of Automation, \\
University of Science and Technology of China \\
\textsuperscript{3}School of Computer Science and Technology, University of Science and Technology of China \\
\texttt{fanjiahe@mail.ustc.edu.cn}
}

\begin{document}
\maketitle
\begin{abstract}
Can large language models with substantially different parameter spaces be merged by direct weighted averaging, without training or semantic alignment? Existing heterogeneous fusion methods typically introduce distillation, adapters, learned latent spaces, routing, or feature alignment, leaving open whether a simpler recipe can work for genuinely different billion-parameter checkpoints. We revisit this counterintuitive question through training-free dimensional adaptation followed by ratio-controlled interpolation. In union-style merging, we expand the smaller model into the larger parameter space; in intersection-style merging, we truncate the larger model into the smaller parameter space (Figure~\ref{fig:merging-strategies}). Across Qwen-family model pairs and benchmarks covering mathematical reasoning, code generation, language understanding, commonsense reasoning, knowledge, and instruction following, deterministic expansion largely preserves the source model function, and small-ratio interpolation can improve over strong source checkpoints by transferring complementary capabilities. However, near-balanced interpolation often collapses, and task-level results reveal a seesaw effect in which gains on some capabilities coexist with regressions on others. These results show that simple parameter averaging, when paired with lightweight dimensional adaptation and carefully controlled ratios, is a surprisingly strong baseline for heterogeneous LLM merging, suggesting that the limits of direct weighted fusion may also bound what more complex heterogeneous merging methods can achieve at scale.
\end{abstract}

\section{Introduction}

\begin{figure}[!t]
\centering
\includegraphics[width=\columnwidth]{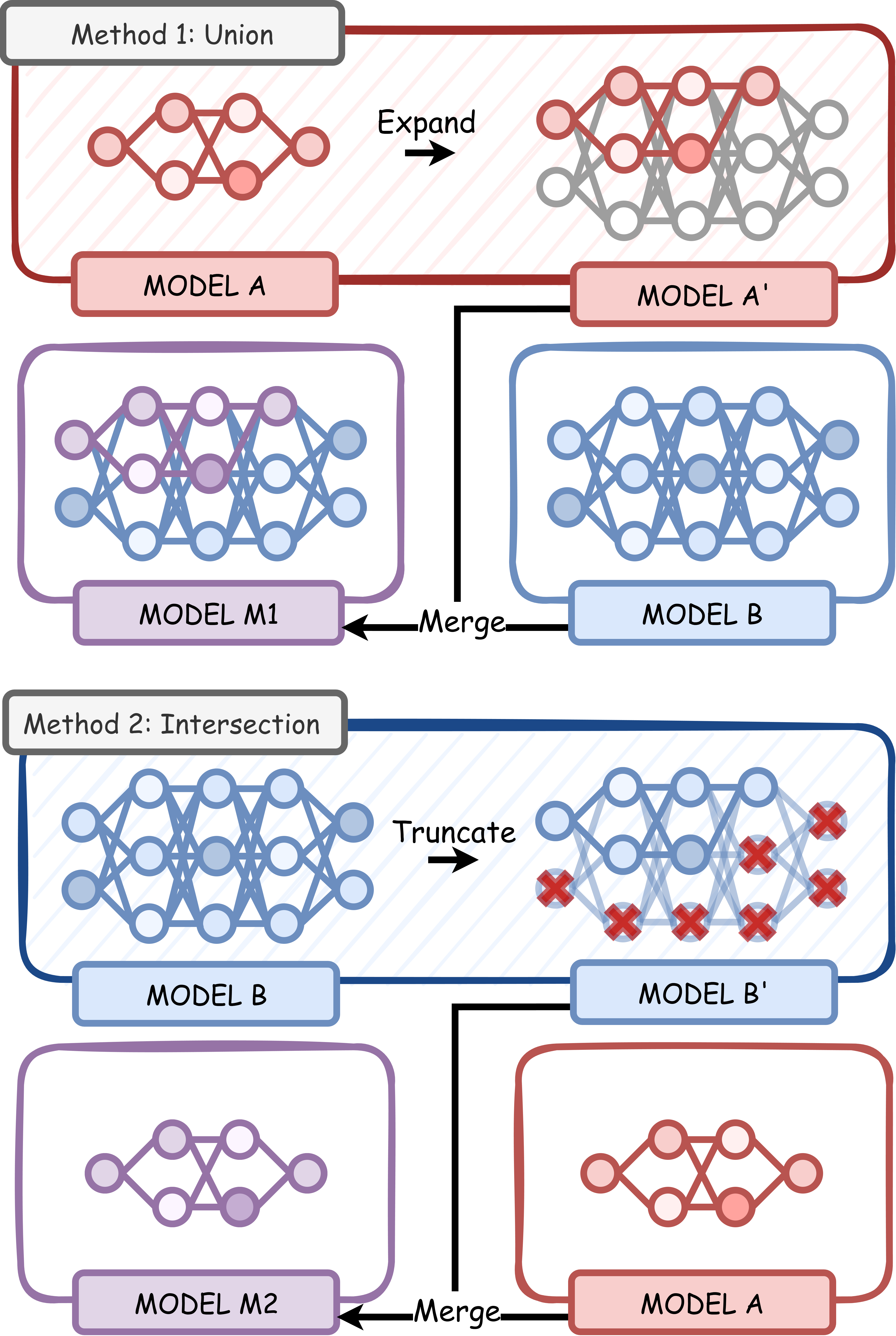}
\caption{Two heterogeneous model merging strategies. (1) Union-style merging: expand the smaller model to match the larger model's dimensionality, then merge. (2) Intersection-style merging: truncate the larger model to the smaller model's dimensionality, then merge.}
\label{fig:merging-strategies}
\end{figure}

Model merging has become a compelling post-training paradigm for large language models, because it offers a direct way to combine capabilities from existing checkpoints without full retraining~\citep{yang2026model,fan2026model,lu2024merge}. In homogeneous settings, a broad family of methods has been proposed, including weight averaging and Model Soups~\citep{wortsman2022model}, task arithmetic~\citep{ilharco2023editing}, TIES-Merging~\citep{yadav2023ties}, Fisher-weighted averaging~\citep{matena2022merging}, RegMean~\citep{jin2023dataless}, and sparsity-based variants~\citep{davari2024breadcrumbs}. Related modular architectures also show the value of selectively routing computation across experts~\citep{jiang2024mixtral}. Yet these successes largely rely on a strong compatibility condition: the source checkpoints share architecture, parameter shapes, and at least partially aligned internal representations. This condition is increasingly restrictive, since real model pools contain checkpoints of different scales, specializations, and design choices~\citep{lu2024merge,chen2025harnessing}.

Heterogeneous model merging relaxes this compatibility condition, but existing solutions often pay for it with extra machinery. Training-based approaches use distillation, generated data, adapters, modular skill storage, or learned latent spaces to transfer knowledge across models~\citep{hinton2015distilling,pfeiffer2021adapterfusion,wan2024knowledgellm,wan2024fusechat,du2025graftllm,soro2026ls}. Training-free approaches reduce mismatch through permutation matching, activation repair, feature zipping, semantic alignment, or coefficient selection~\citep{ainsworth2023git,jordan2023repair,stoica2024zipit,gu2025seme,du2025adamms,hackmann2024hm3}. These methods are valuable, but they leave a sharper question unresolved: can we merge sufficiently different billion-parameter language models with low cost, no additional training, and no explicit semantic alignment? This question is especially important because direct weighted averaging across mismatched parameter spaces appears counterintuitive: a larger model and a smaller model do not even expose tensors of the same shape, and representational mismatch weakens the assumptions behind neuron-wise or feature-wise correspondence.

The technical challenge has two parts. First, parameter-shape mismatch prevents direct interpolation, so the models must be placed in a common parameter space before any weighted averaging is possible. Second, even after shape compatibility is enforced, interpolation can destroy capability if the two solutions are not connected by a stable low-loss path; related homogeneous studies show that interpolation depends strongly on mode connectivity, permutation alignment, and representation similarity~\citep{garipov2018loss,frankle2020linear,ainsworth2023git,raghu2017svcca,kornblith2019similarity}. Recent work on cross-scale fusion and scaling behavior further suggests that the effectiveness of merging rules may change at LLM scale~\citep{chen2026heterofusion,wang2025mergingscaling}. This motivates a controlled test of the simplest possible hypothesis: after deterministic dimensional adaptation, is direct weighted averaging actually enough, provided that the mixing ratio is small?

We answer this question through training-free dimensional adaptation followed by ratio-controlled interpolation. In the union-style setting, we expand the smaller checkpoint into the larger parameter space before merging; in the intersection-style setting, we truncate the larger checkpoint into the smaller parameter space before merging. The resulting procedure uses no fine-tuning, no distillation data, no adapters, no routing module, no learned latent encoder, and no semantic alignment step. This makes the intervention deliberately simple and interpretable: performance changes can be attributed primarily to dimensional adaptation, mixing ratio, and the compatibility of the two source checkpoints.

Across Qwen-family model pairs at billion-parameter scale and diverse benchmark families, this simple procedure works more often than the standard intuition would suggest. Deterministic expansion largely preserves the smaller model's function, showing that dimensional adaptation can act as a functional bridge rather than only a tensor-shape repair. Small-ratio interpolation can further improve performance by transferring complementary capabilities; for example, in union-style merging between Qwen2.5-14B and Qwen2.5-32B, the two source checkpoints obtain average scores of 0.7044 and 0.7424, respectively, while the best merged model reaches 0.7525, and intersection-style merging improves Qwen2.5-3B from 0.5459 to 0.5716 when a truncated 32B branch is injected with a small coefficient. At the same time, the method has clear limits: near-balanced interpolation often collapses, and task-level results reveal a seesaw effect where improvements on some capabilities coexist with regressions on others.

These findings suggest that heterogeneous LLM merging may follow a trajectory similar to homogeneous merging at scale: simple parameter averaging, when used in the right effective region, can be a surprisingly strong baseline, and the regimes where it fails may expose fundamental compatibility bottlenecks that more complex methods cannot easily bypass. Our contributions are fourfold: (1) we formulate low-cost, training-free, alignment-free heterogeneous LLM merging under substantial parameter-scale mismatch; (2) we introduce union-style and intersection-style dimensional adaptation as simple probes for direct weighted fusion; (3) we identify ratio-sensitive regimes of preservation, transfer, and collapse across billion-parameter model pairs; and (4) we show that task-level complementarity and seesaw behavior are central diagnostics for understanding when heterogeneous merging improves capability and when it merely redistributes or destroys it.

\section{Related Work}

\begin{figure*}[!t]
\centering
\includegraphics[width=0.98\textwidth]{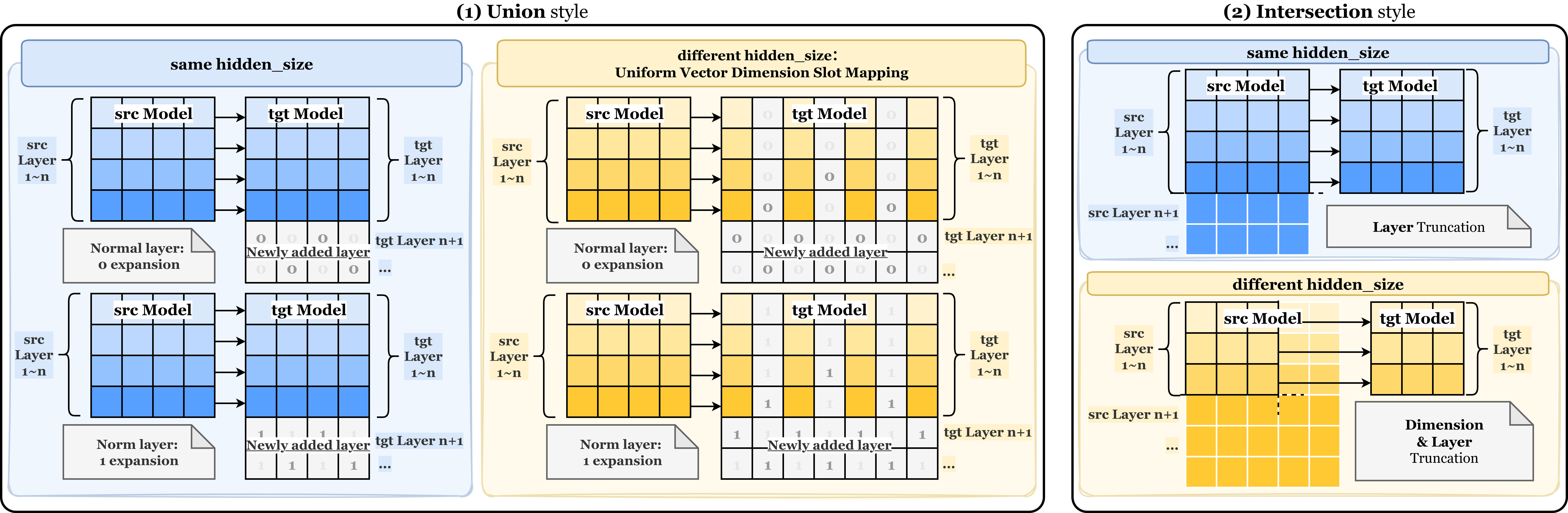}
\caption{Dimensional adaptation procedures for heterogeneous model merging. (a) Initialization scheme for heterogeneous layer extension. Compatible source tensors are copied into the target model; newly introduced attention and MLP branches are initialized to zero, and normalization scales are initialized to one. (b) Truncation procedure for intersection-style merging. A larger checkpoint is projected into the smaller parameter space by copying compatible tensors, slicing oversized tensors, and removing layers beyond the target depth.}
\label{fig:dimensional-adaptation}
\end{figure*}

Recent surveys summarize model merging for LLMs, MLLMs, and related foundation-model settings~\citep{yang2026model,fan2026model}.

\paragraph{Homogeneous model merging.}
Prior work has largely focused on homogeneous settings, where models share architecture and parameter layout. Weight-based methods include weight averaging and Model Soups~\citep{wortsman2022model}, task arithmetic~\citep{ilharco2023editing}, TIES-Merging~\citep{yadav2023ties}, and DARE-style variants~\citep{yu2024language}. Geometry- or statistics-based methods include SLERP~\citep{shoemake1985animating}, Fisher-weighted averaging~\citep{matena2022merging}, and RegMean~\citep{jin2023dataless}. Sparse-mask merging reduces interference by retaining selected parameter updates~\citep{davari2024breadcrumbs}, while MoE-style routing offers a related modular route for activating specialized experts~\citep{jiang2024mixtral}.

\paragraph{Limitations in homogeneous settings.}
Homogeneous merging relies on aligned parameter shapes, similar internal representations, and often careful hyperparameter tuning. More sophisticated strategies add optimization or engineering overhead, and scaling-law evidence suggests that merging gains do not automatically scale with model size~\citep{wang2025mergingscaling}. In a recent in-the-wild evaluation with heterogeneous, potentially overlapping or conflicting fine-tuned checkpoints, Task Arithmetic was the only evaluated method that reliably yielded gains, while more complex subspace- or interference-aware methods often failed to provide clear improvements~\citep{hitit2026systematic}. These constraints limit reuse of diverse real-world model pools.

\paragraph{Subspace-based merging and benchmarks.}
A closely related line reduces interference by localizing or matching task-relevant parameter subspaces. TALL-masks localize sparse task information and use the resulting supports for compression and consensus merging~\citep{wang2024localizing}; Localize-and-Stitch identifies compact task-localized regions and stitches them into a pretrained model through sparse task arithmetic~\citep{he2024localize}; and MaTS formulates merging as matching models in task parameter subspaces through a linear-system solution~\citep{tam2024merging}. These methods are connected to our intersection-style setting because both restrict the region of parameter space that participates in merging. The difference is that subspace-based methods usually estimate task supports or matching subspaces for shape-compatible task vectors, whereas our truncate-then-merge protocol first applies a deterministic architecture projection from a larger checkpoint to a smaller parameter space and then performs small-ratio interpolation. We therefore ask whether useful projected perturbations can be absorbed without learning a task subspace, solving a matching problem, or requiring additional optimization. Recent benchmarks such as MergeBench further emphasize the need for explicit evaluation of domain-specialized LLM merging methods and failure modes~\citep{he2025mergebench}; our setting is complementary because it focuses on cross-scale dimensional adaptation rather than standard same-shape merging.

\paragraph{Heterogeneous model merging.}
Heterogeneous merging is more challenging because parameter-wise averaging is no longer obviously meaningful when source models differ in width, depth, module layout, or even modality. Existing approaches therefore often introduce additional mechanisms to make cross-model reuse feasible. One line of work uses training or learned intermediate modules: knowledge distillation transfers behavior through teacher--student optimization~\citep{hinton2015distilling}, AdapterFusion composes trained adapters rather than raw checkpoints~\citep{pfeiffer2021adapterfusion}, FuseLLM and FuseChat fuse knowledge through model-generated data and subsequent training~\citep{wan2024knowledgellm,wan2024fusechat}, GraftLLM stores capabilities in modular SkillPacks~\citep{du2025graftllm}, and LS-Merge learns a latent weight space before merging language models~\citep{soro2026ls}. These methods broaden the reuse of heterogeneous model assets, but their reliance on optimization, generated or curated data, trained modules, or latent encoders weakens the original appeal of model merging as a fast training-free operation.

A second line seeks training-free compatibility through alignment or repair. Git Re-Basin aligns models modulo permutation symmetries~\citep{ainsworth2023git}, REPAIR corrects activation statistics after permutation-based interpolation~\citep{jordan2023repair}, and ZipIt merges models by matching and zipping related features across networks~\citep{stoica2024zipit}. For language and multimodal models, semantic-alignment and coefficient-selection methods further reduce mismatch before interpolation, as in SeMe and AdaMMS~\citep{gu2025seme,du2025adamms}, while HM3 studies heterogeneous multi-class classifier merging~\citep{hackmann2024hm3}. However, many of these methods either focus on relatively small networks, require explicit semantic or feature alignment, or study heterogeneity that remains close to a shared base or task structure. As a result, it remains unclear whether sufficiently different billion-parameter language models can be merged by a simpler recipe: adapt dimensions without training and then directly interpolate weights.

Recent studies on cross-scale language-model fusion and model-merging scaling further motivate this question~\citep{chen2026heterofusion,wang2025mergingscaling}. In particular, if sophisticated merging rules become less distinguishable at larger scale~\citep{yadav2025what}, heterogeneous LLM merging may follow a similar trajectory: carefully controlled simple weighted interpolation could be a stronger baseline than expected. Our work tests this possibility directly by removing learned alignment, adapters, routing, and post-hoc training, and by asking whether dimensional adaptation plus small-ratio interpolation can preserve or improve capabilities across genuinely different LLM scales.

\paragraph{Position of this work.}
We target training-free heterogeneous merging under parameter-scale mismatch with lightweight dimensional adaptation and ratio-controlled interpolation, preserving simple merging while extending it to broader model pairs.

\section{Method}

\subsection{Problem Setup}

We consider two pretrained models, denoted by $A$ and $B$, where model $A$ is no larger than model $B$ in parameter dimensionality. Let $\theta_A \in \mathbb{R}^{d_A}$ and $\theta_B \in \mathbb{R}^{d_B}$ denote their parameter vectors, with $d_A \le d_B$. Let $\mathcal{X}$ denote the input space, $\mathcal{D}$ the evaluation distribution, and $\ell$ the task loss. We study two complementary merging settings, referred to as \emph{union-style merging} and \emph{intersection-style merging}, to characterize when heterogeneous merging can preserve capability and when it fails.

\subsection{Union-Style Merging (Expand Then Merge)}
We first map model $A$ to the larger space of model $B$ through an expansion operator
\begin{equation}
\mathcal{E}: \mathbb{R}^{d_A} \rightarrow \mathbb{R}^{d_B}, \qquad \tilde{\theta}_A = \mathcal{E}(\theta_A).
\end{equation}
Expansion deterministically copies compatible source tensors into the larger target tensors and initializes newly introduced dimensions to minimize functional perturbation, as illustrated in Figure~\ref{fig:dimensional-adaptation}(a): newly introduced linear parameters are zero-initialized, normalization scales are initialized to neutral values, and additional Transformer blocks~\citep{vaswani2017attention} are initialized as residual identity mappings. Architecture-aware mappings preserve grouped-query attention structure~\citep{ainslie2023gqa} and expand MLP matrices along their semantic input or output dimensions. Further normalization details are given in Appendix~\ref{app:method-details}. We then perform convex interpolation in the high-dimensional space:
\begin{equation}
\theta^{\cup}(\lambda) = (1-\lambda)\tilde{\theta}_A + \lambda\theta_B, \qquad \lambda \in [0,1].
\end{equation}

\subsection{Intersection-Style Merging (Truncate Then Merge)}
Complementary to expansion, we study an intersection-style setting that projects the larger model $B$ into the smaller parameter space of model $A$ via a truncation operator:
\begin{equation}
\mathcal{T}: \mathbb{R}^{d_B} \rightarrow \mathbb{R}^{d_A}, \qquad \tilde{\theta}_B = \mathcal{T}(\theta_B).
\end{equation}
As shown in Figure~\ref{fig:dimensional-adaptation}(b), we instantiate the small-model configuration and initialize it from the large checkpoint by copying shape-compatible tensors, slicing oversized tensors, and removing layers beyond the target depth. Unlike expansion, truncation discards parameters and is inherently lossy, so the truncated branch must remain a small perturbation. Merging is performed in the low-dimensional space:
\begin{equation}
\theta^{\cap}(\mu) = (1-\mu)\theta_A + \mu\tilde{\theta}_B, \qquad \mu \in [0,1].
\end{equation}

\subsection{Evaluation Protocol}
Union-style and intersection-style merging are evaluated as two separate experimental protocols. For each protocol, we search over its own ratio grid and select the best checkpoint by downstream validation performance:
\begin{equation}
\theta_{\cup}^{*}
=
\arg\min_{\lambda\in\Lambda}
\mathcal{L}_{\mathrm{eval}}\big(\theta^{\cup}(\lambda)\big),
\end{equation}
\begin{equation}
\theta_{\cap}^{*}
=
\arg\min_{\mu\in M}
\mathcal{L}_{\mathrm{eval}}\big(\theta^{\cap}(\mu)\big),
\end{equation}
where the corresponding ratio grids are specified in the Experiment section. The key empirical question is whether useful heterogeneous transfer appears in small-ratio neighborhoods of either anchor model, and where interpolation collapses within each protocol.

\section{Experiment}

\begin{table}[!t]
\centering
\small
\begin{tabularx}{\columnwidth}{>{\raggedright\arraybackslash}X>{\raggedright\arraybackslash}X}
\hline
  \textbf{Family} & \textbf{Benchmarks} \\
\hline
Math & GSM8K, MATH-500 \\
Code generation & HumanEvalPlus, MBPPPlus \\
Instruction following & IFEval \\
QA and NLU & MNLI, RTE, QNLI, PIQA, Winogrande, COPA, BoolQ \\
Commonsense \& knowledge & ARC, HellaSwag, MMLU \\
Comprehensive reasoning & BBH \\
\hline
\end{tabularx}
\caption{Benchmark families used in our experiments.}
\label{tab:exp-datasets}
\end{table}

\begin{table}[!t]
\centering
\small
\begin{tabularx}{\columnwidth}{@{}p{0.12\columnwidth}X X@{}}
\toprule
\textbf{ID} & \textbf{Smaller model} & \textbf{Larger model} \\
\midrule
P1 & Qwen2.5-14B & Qwen2.5-32B \\
P2 & Qwen2.5-3B & Qwen2.5-32B \\
P3 & Qwen3-4B-Thinking-2507~\citep{qwen3thinking2507} & Qwen3-8B \\
P4 & Qwen3-14B & Qwen3-32B \\
\bottomrule
\end{tabularx}
\caption{Model-pair configuration for heterogeneous merging.}
\label{tab:exp-model-pairs}
\end{table}

\begin{table}[!t]
\centering
\small
\setlength{\tabcolsep}{3pt}
\renewcommand{\arraystretch}{1.05}
\begin{tabularx}{0.95\columnwidth}{@{}>{\raggedright\arraybackslash}p{0.28\columnwidth}*{2}{>{\centering\arraybackslash}X}>{\raggedleft\arraybackslash}X@{}}
\toprule
\textbf{Dataset} & \textbf{Base A} & \textbf{Base B} & \textbf{Merged} \\
\midrule
\texttt{gsm8k} & 0.8779 & 0.9014 & \textbf{0.9045} \\
\texttt{MATH-500} & 0.5460 & \textbf{0.6180} & 0.6140 \\
\texttt{humanevalplus} & 0.5604 & 0.6165 & \mergedgain{0.6494}{0.03} \\
\texttt{mbppplus} & 0.6738 & 0.7405 & \textbf{0.7413} \\
\texttt{BoolQ} & 0.8532 & \textbf{0.8740} & 0.8719 \\
\texttt{COPA} & 0.9000 & 0.8700 & \mergedgain{0.9100}{0.01} \\
\texttt{MNLI} & 0.6689 & 0.7217 & \textbf{0.7300} \\
\texttt{PIQA} & 0.8248 & 0.8243 & \textbf{0.8254} \\
\texttt{QNLI} & 0.6771 & \textbf{0.8596} & 0.8594 \\
\texttt{RTE} & 0.7906 & \textbf{0.8159} & 0.8123 \\
\texttt{Winogrande} & 0.7569 & 0.7569 & \mergedgain{0.7774}{0.02} \\
\texttt{ARC} & 0.5241 & 0.5320 & \textbf{0.5341} \\
\texttt{HellaSwag} & 0.6911 & \textbf{0.7175} & 0.7156 \\
\texttt{MMLU} & 0.7094 & 0.7620 & \textbf{0.7630} \\
\texttt{ifeval} & 0.4424 & 0.4832 & \mergedgain{0.5072}{0.02} \\
\texttt{BBH} & 0.7731 & 0.7852 & \mergedgain{0.8241}{0.04} \\
\midrule
\textbf{Avg.} & 0.7044 & 0.7424 & \mergedgain{0.7525}{0.01} \\
\bottomrule
\end{tabularx}
\caption{Union merging results for the Qwen2.5-14B and Qwen2.5-32B pair. The merged checkpoint combines Qwen2.5-32B with the expanded Qwen2.5-14B model.}
\label{tab:union-qwen25}
\end{table}

\begin{table}[!t]
\centering
\small
\setlength{\tabcolsep}{3pt}
\renewcommand{\arraystretch}{1.05}
\begin{tabularx}{0.95\columnwidth}{@{}>{\raggedright\arraybackslash}p{0.28\columnwidth}*{2}{>{\centering\arraybackslash}X}>{\raggedleft\arraybackslash}X@{}}
\toprule
\textbf{Dataset} & \textbf{Base A} & \textbf{Base B} & \textbf{Merged} \\
\midrule
\texttt{gsm8k} & 0.8931 & 0.8764 & \textbf{0.9007} \\
\texttt{MATH-500} & 0.5100 & 0.5300 & \mergedgain{0.5480}{0.02} \\
\texttt{humanevalplus} & 0.5854 & 0.6091 & \mergedgain{0.6768}{0.07} \\
\texttt{mbppplus} & 0.4550 & 0.6328 & \textbf{0.6402} \\
\texttt{BoolQ} & 0.7477 & 0.8673 & \textbf{0.8709} \\
\texttt{COPA} & 0.8200 & 0.8500 & \mergedgain{0.8700}{0.02} \\
\texttt{MNLI} & 0.5849 & 0.6220 & \mergedgain{0.6387}{0.02} \\
\texttt{PIQA} & 0.7546 & 0.7753 & \textbf{0.7775} \\
\texttt{QNLI} & 0.5266 & 0.7824 & \textbf{0.7899} \\
\texttt{RTE} & \textbf{0.8123} & 0.7762 & \textbf{0.8123} \\
\texttt{Winogrande} & 0.6590 & 0.6765 & \textbf{0.6851} \\
\texttt{ARC} & 0.3887 & \textbf{0.4640} & 0.4610 \\
\texttt{HellaSwag} & 0.5152 & \textbf{0.6125} & 0.6094 \\
\texttt{MMLU} & 0.5854 & 0.6695 & \textbf{0.6722} \\
\texttt{ifeval} & 0.4652 & 0.4233 & \mergedgain{0.4880}{0.02} \\
\texttt{BBH} & 0.7380 & \textbf{0.7889} & 0.7852 \\
\midrule
\textbf{Avg.} & 0.6276 & 0.6848 & \mergedgain{0.7016}{0.02} \\
\bottomrule
\end{tabularx}
\caption{Union merging results for the Qwen3-4B-Thinking-2507 and Qwen3-8B pair. The merged checkpoint combines Qwen3-8B with the expanded Qwen3-4B-Thinking-2507 model.}
\label{tab:union-qwen3-4b}
\end{table}

\begin{figure}[!t]
\centering
\includegraphics[width=0.92\columnwidth]{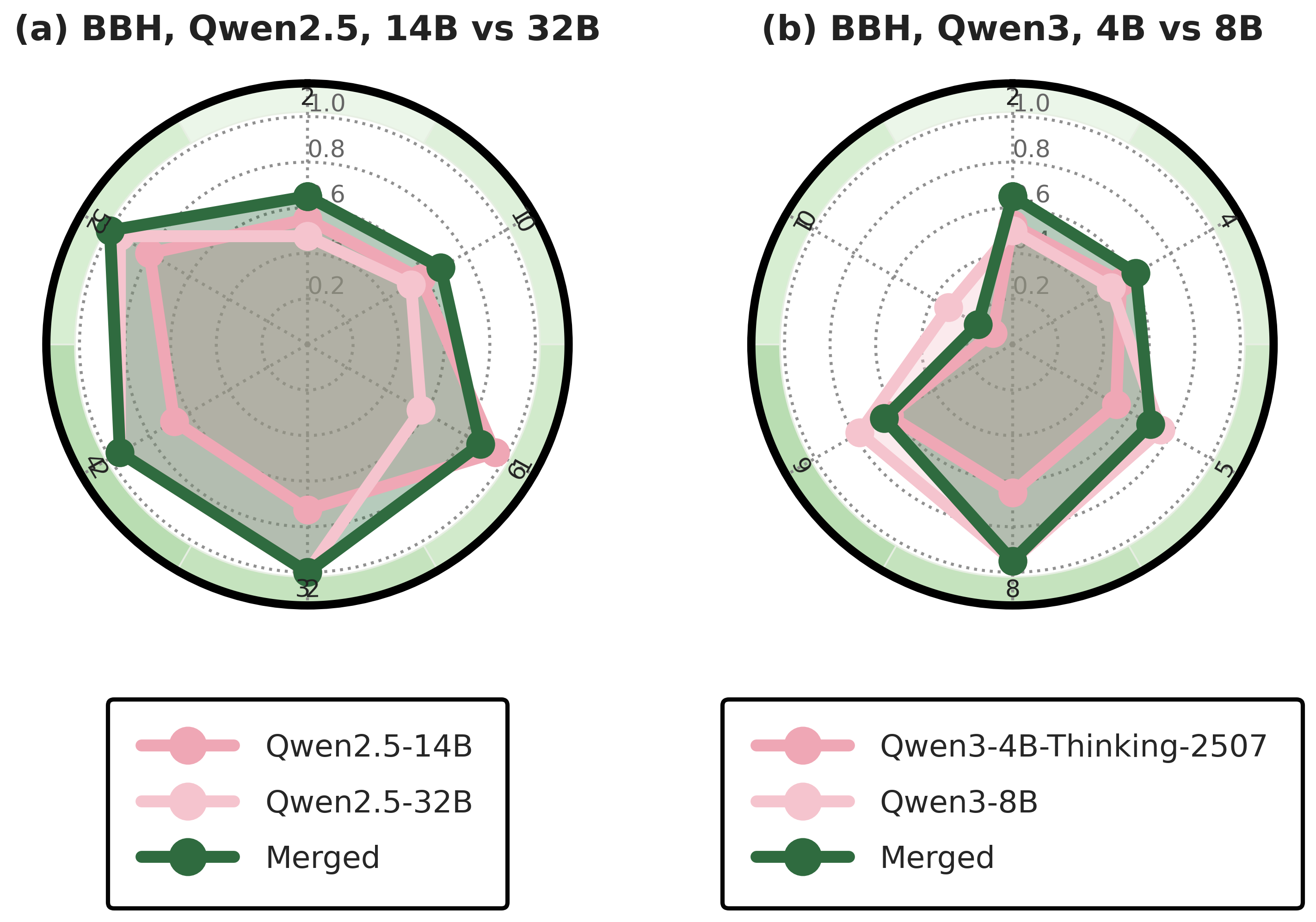}
\caption{Knowledge-structure and merging-effect radar comparison. Intersecting but non-nested competence profiles leave complementary regions, and the merged model improves by about four percentage points over both sources; nested profiles offer limited new information, and accuracy changes only marginally.}
\label{fig:radar-knowledge-structure}
\end{figure}

\begin{table}[!t]
\centering
\small
\setlength{\tabcolsep}{3pt}
\renewcommand{\arraystretch}{1.05}
\begin{tabularx}{0.95\columnwidth}{@{}>{\raggedright\arraybackslash}p{0.28\columnwidth}*{2}{>{\centering\arraybackslash}X}>{\raggedleft\arraybackslash}X@{}}
\toprule
\textbf{Dataset} & \textbf{Base A} & \textbf{Base B} & \textbf{Merged} \\
\midrule
\texttt{gsm8k} & \textbf{0.9227} & 0.9113 & 0.9181 \\
\texttt{MATH-500} & 0.8160 & 0.7740 & \textbf{0.8180} \\
\texttt{humanevalplus} & 0.6707 & 0.6890 & \mergedgain{0.7378}{0.05} \\
\texttt{mbppplus} & 0.6878 & 0.6402 & \mergedgain{0.7169}{0.03} \\
\texttt{BoolQ} & \textbf{0.8924} & 0.8633 & 0.8896 \\
\texttt{COPA} & 0.9000 & 0.9400 & \mergedgain{0.9600}{0.02} \\
\texttt{MNLI} & 0.6706 & 0.6989 & \mergedgain{0.7104}{0.01} \\
\texttt{PIQA} & 0.7976 & 0.8150 & \textbf{0.8188} \\
\texttt{QNLI} & 0.8380 & 0.8104 & \mergedgain{0.8543}{0.02} \\
\texttt{RTE} & 0.7762 & 0.7545 & \mergedgain{0.7906}{0.01} \\
\texttt{Winogrande} & 0.7277 & \textbf{0.7301} & 0.7261 \\
\texttt{ARC} & 0.5093 & 0.5341 & \textbf{0.5352} \\
\texttt{HellaSwag} & 0.6619 & \textbf{0.7065} & 0.7061 \\
\texttt{MMLU} & 0.7135 & \textbf{0.7621} & 0.7604 \\
\texttt{ifeval} & 0.4484 & 0.4652 & \mergedgain{0.5635}{0.10} \\
\texttt{BBH} & 0.8500 & 0.8639 & \mergedgain{0.8843}{0.02} \\
\midrule
\textbf{Avg.} & 0.7427 & 0.7474 & \mergedgain{0.7744}{0.03} \\
\bottomrule
\end{tabularx}
\caption{Union merging results for the Qwen3-14B and Qwen3-32B pair. The merged checkpoint combines Qwen3-32B with the expanded Qwen3-14B model.}
\label{tab:union-qwen3-14b}
\end{table}
We evaluate Qwen2.5~\citep{qwen25} and Qwen3~\citep{qwen3} model pairs ranging from 3B to 32B parameters. The benchmark suite spans mathematical reasoning, code generation, instruction following, NLU, commonsense, knowledge, and comprehensive reasoning, using GSM8K~\citep{cobbe2021training}, MATH~\citep{hendrycks2021measuring,lightman2023lets}, HumanEval/EvalPlus~\citep{chen2021evaluating,evalplus2023}, IFEval~\citep{zhou2023instruction}, GLUE-style NLU tasks~\citep{williams2018broad,wang2018glue}, PIQA~\citep{bisk2020piqa}, Winogrande~\citep{sakaguchi2020winogrande}, COPA~\citep{roemmele2011choice}, BoolQ~\citep{clark2019boolq}, ARC~\citep{clark2018think}, HellaSwag~\citep{zellers2019hellaswag}, MMLU~\citep{hendrycks2020measuring}, and BBH~\citep{suzgun2022challenging}. We use the EleutherAI LM Evaluation Harness~\citep{eval-harness} and BigCode Evaluation Harness~\citep{bigcode-eval-harness}. The benchmark families and model-pair configurations are summarized in Tables~\ref{tab:exp-datasets} and~\ref{tab:exp-model-pairs}.

\begin{table}[!t]
\centering
\small
\setlength{\tabcolsep}{4pt}
\renewcommand{\arraystretch}{1.05}
\begin{tabularx}{\columnwidth}{@{}>{\raggedright\arraybackslash}X
                                >{\centering\arraybackslash}p{0.25\columnwidth}
                                >{\raggedleft\arraybackslash}p{0.18\columnwidth}@{}}
\toprule
\textbf{Dataset} & \textbf{Qwen2.5-3B} & \textbf{Merged} \\
\midrule
\texttt{gsm8k} & 0.6566 & \mergedgain{0.7210}{0.06} \\
\texttt{MATH-500} & 0.3840 & \mergedgain{0.4060}{0.02} \\
\texttt{humanevalplus} & \textbf{0.4390} & 0.4329 \\
\texttt{mbppplus} & 0.5899 & \mergedgain{0.6058}{0.02} \\
\texttt{BoolQ} & 0.7080 & \mergedgain{0.7272}{0.02} \\
\texttt{COPA} & 0.7800 & \mergedgain{0.8000}{0.02} \\
\texttt{MNLI} & 0.4598 & \mergedgain{0.5334}{0.07} \\
\texttt{PIQA} & 0.7579 & \textbf{0.7584} \\
\texttt{QNLI} & 0.5286 & \mergedgain{0.5526}{0.02} \\
\texttt{RTE} & 0.6426 & \mergedgain{0.7726}{0.13} \\
\texttt{Winogrande} & 0.6338 & \textbf{0.6401} \\
\texttt{ARC} & \textbf{0.3442} & 0.3430 \\
\texttt{HellaSwag} & \textbf{0.5094} & 0.5087 \\
\texttt{MMLU} & 0.4496 & \textbf{0.4514} \\
\texttt{ifeval} & 0.4041 & \textbf{0.4113} \\
\texttt{BBH} & 0.4463 & \mergedgain{0.4806}{0.03} \\
\midrule
\textbf{Avg.} & 0.5459 & \mergedgain{0.5716}{0.03} \\
\bottomrule
\end{tabularx}
\caption{Intersection merging results for Qwen2.5-3B and a Qwen2.5-32B checkpoint truncated to 3B size.}
\label{tab:intersection-qwen25-3b}
\end{table}

\begin{table}[!t]
\centering
\small
\setlength{\tabcolsep}{4pt}
\renewcommand{\arraystretch}{1.05}
\begin{tabularx}{\columnwidth}{@{}>{\raggedright\arraybackslash}X
                                >{\centering\arraybackslash}p{0.28\columnwidth}
                                >{\raggedleft\arraybackslash}p{0.14\columnwidth}@{}}
\toprule
\textbf{Dataset} & \textbf{\mbox{Qwen2.5-14B}} & \textbf{Merged} \\
\midrule
\texttt{gsm8k} & 0.8802 & \textbf{0.8848} \\
\texttt{MATH-500} & 0.5460 & \mergedgain{0.5640}{0.02} \\
\texttt{humanevalplus} & 0.5604 & \textbf{0.5616} \\
\texttt{mbppplus} & 0.6738 & \textbf{0.6775} \\
\texttt{BoolQ} & 0.8532 & \textbf{0.8612} \\
\texttt{COPA} & 0.9000 & \mergedgain{0.9100}{0.01} \\
\texttt{MNLI} & 0.6689 & \mergedgain{0.6880}{0.02} \\
\texttt{PIQA} & 0.8248 & \textbf{0.8254} \\
\texttt{QNLI} & 0.6771 & \mergedgain{0.7657}{0.09} \\
\texttt{RTE} & 0.7906 & \textbf{0.7978} \\
\texttt{Winogrande} & 0.7569 & \mergedgain{0.7774}{0.02} \\
\texttt{ARC} & \textbf{0.5241} & 0.5200 \\
\texttt{HellaSwag} & \textbf{0.6911} & 0.6903 \\
\texttt{MMLU} & \textbf{0.7094} & 0.7088 \\
\texttt{ifeval} & \textbf{0.4424} & 0.4388 \\
\texttt{BBH} & 0.7731 & \mergedgain{0.7870}{0.01} \\
\midrule
\textbf{Avg.} & 0.7045 & \mergedgain{0.7161}{0.01} \\
\bottomrule
\end{tabularx}
\caption{Intersection merging results for Qwen2.5-14B and a Qwen2.5-32B checkpoint truncated to 14B size.}
\label{tab:intersection-qwen25-14b}
\end{table}

\begin{table}[!t]
\centering
\small
\setlength{\tabcolsep}{4pt}
\renewcommand{\arraystretch}{1.05}
\begin{tabularx}{\columnwidth}{@{}>{\raggedright\arraybackslash}X
                                >{\centering\arraybackslash}p{0.28\columnwidth}
                                >{\raggedleft\arraybackslash}p{0.14\columnwidth}@{}}
\toprule
\textbf{Dataset} & \textbf{\mbox{Qwen3-14B}} & \textbf{Merged} \\
\midrule
\texttt{gsm8k} & 0.9340 & \textbf{0.9386} \\
\texttt{MATH-500} & 0.8160 & \mergedgain{0.8280}{0.01} \\
\texttt{humanevalplus} & 0.7500 & \mergedgain{0.7683}{0.02} \\
\texttt{mbppplus} & 0.7381 & \textbf{0.7434} \\
\texttt{BoolQ} & 0.8924 & \textbf{0.8933} \\
\texttt{COPA} & \textbf{0.9000} & \textbf{0.9000} \\
\texttt{MNLI} & \textbf{0.6706} & 0.6699 \\
\texttt{PIQA} & 0.7976 & \textbf{0.8025} \\
\texttt{QNLI} & 0.8380 & \textbf{0.8446} \\
\texttt{RTE} & 0.7762 & \mergedgain{0.7978}{0.02} \\
\texttt{Winogrande} & 0.7277 & \textbf{0.7340} \\
\texttt{ARC} & 0.5093 & \textbf{0.5127} \\
\texttt{HellaSwag} & 0.6619 & \textbf{0.6683} \\
\texttt{MMLU} & \textbf{0.7135} & 0.7129 \\
\texttt{ifeval} & 0.4484 & \mergedgain{0.4712}{0.02} \\
\texttt{BBH} & 0.8500 & \mergedgain{0.8602}{0.01} \\
\midrule
\textbf{Avg.} & 0.7515 & \mergedgain{0.7591}{0.01} \\
\bottomrule
\end{tabularx}
\caption{Intersection merging results for Qwen3-14B and a Qwen3-32B checkpoint truncated to 14B size.}
\label{tab:intersection-qwen3-14b}
\end{table}

\begin{table}[!t]
\centering
\small
\setlength{\tabcolsep}{4pt}
\renewcommand{\arraystretch}{1.05}
\begin{tabularx}{\columnwidth}{@{}>{\raggedright\arraybackslash}X
                                >{\centering\arraybackslash}p{0.40\columnwidth}
                                >{\raggedleft\arraybackslash}p{0.14\columnwidth}@{}}
\toprule
\textbf{Dataset} & \textbf{Qwen3-4B} & \textbf{Merged} \\
\midrule
\texttt{gsm8k} & \textbf{0.8863} & 0.8840 \\
\texttt{MATH-500} & 0.5100 & \mergedgain{0.5380}{0.03} \\
\texttt{humanevalplus} & \textbf{0.5220} & 0.5055 \\
\texttt{mbppplus} & \textbf{0.5275} & 0.5193 \\
\texttt{BoolQ} & 0.7477 & \mergedgain{0.7798}{0.03} \\
\texttt{COPA} & \textbf{0.8200} & 0.8100 \\
\texttt{MNLI} & 0.5849 & \textbf{0.5887} \\
\texttt{PIQA} & 0.7546 & \textbf{0.7563} \\
\texttt{QNLI} & 0.5266 & \mergedgain{0.5863}{0.06} \\
\texttt{RTE} & \textbf{0.8123} & 0.8087 \\
\texttt{Winogrande} & 0.6590 & \textbf{0.6748} \\
\texttt{ARC} & \textbf{0.3887} & 0.3872 \\
\texttt{HellaSwag} & \textbf{0.5152} & 0.5146 \\
\texttt{MMLU} & 0.5854 & \textbf{0.5940} \\
\texttt{ifeval} & 0.4652 & \mergedgain{0.4772}{0.01} \\
\texttt{BBH} & 0.7380 & \mergedgain{0.7657}{0.03} \\
\midrule
\textbf{Avg.} & 0.6277 & \mergedgain{0.6369}{0.01} \\
\bottomrule
\end{tabularx}
\caption{Intersection merging results for Qwen3-4B-Thinking-2507 and a Qwen3-8B checkpoint truncated to 4B size.}
\label{tab:intersection-qwen3-4b}
\end{table}

Let $\theta_s$ and $\theta_l$ denote the smaller and larger checkpoints, and let $\Pi_{s\to l}$ and $\Pi_{l\to s}$ denote expansion and truncation operators. We evaluate:
\begin{equation}
\theta_{\mathrm{union}}(\lambda)=(1-\lambda)\Pi_{s\to l}(\theta_s)+\lambda\theta_l,
\end{equation}
\begin{equation}
\theta_{\mathrm{inter}}(\mu)=(1-\mu)\theta_s+\mu\Pi_{l\to s}(\theta_l),
\end{equation}
We use fixed ratio grids for the two protocols. For union-style merging, $\lambda \in \{0.02, 0.04, 0.96, 0.98\}$ controls the contribution of the larger checkpoint. For intersection-style merging, $\mu \in \{0.01, 0.02, 0.03, 0.04, 0.05\}$ controls the contribution of the truncated larger checkpoint. The merged checkpoints are evaluated on the benchmark families reported for each protocol.

\section{Main Results}

Across three union-style settings, direct weighted averaging after deterministic expansion consistently improves the average score over both source checkpoints (Tables~\ref{tab:union-qwen25}--\ref{tab:union-qwen3-14b}). Intersection-style merging also yields positive gains when the truncated larger checkpoint is injected with a small coefficient: Qwen2.5-3B improves from 0.5459 to 0.5716, and Qwen2.5-14B improves from 0.7045 to 0.7161 (Tables~\ref{tab:intersection-qwen25-3b} and~\ref{tab:intersection-qwen25-14b}). The gains are largest when the two sources provide complementary competence rather than near-nested capabilities, supporting the view that heterogeneous merging can transfer useful signal without training or explicit alignment.

\subsection{Knowledge-Structure Compatibility and Complementarity}

The experimental results indicate that heterogeneous merging can produce meaningful complementarity when the two source models encode partially distinct competence profiles. This profile-based view is related in spirit to representation-similarity analyses that compare learned structures across neural models~\citep{raghu2017svcca,kornblith2019similarity}. At the aggregate level, the best merged configuration can outperform both source checkpoints, and task-level results further show that gains are distributed across multiple categories rather than concentrated in a single outlier task. This supports a practical interpretation of ``complementary transfer'': when knowledge structure is sufficiently different, merging can inherit strengths from both sides and realize a clear compensation effect.

Figure~\ref{fig:radar-knowledge-structure} clarifies the structural condition behind this behavior. When the larger model nearly encloses the smaller model, overlap dominates and merging changes accuracy only marginally. When the two models exhibit an intersecting but non-nested relationship, merging introduces complementary signals and yields a visible gain of about four percentage points.

\subsection{Scale Gap Drives Intersection Gains}

We further evaluate intersection-style merging by merging Qwen2.5-3B with a Qwen2.5-32B checkpoint truncated to the 3B parameter space. The merged model is computed as $(1-\mu)$Qwen2.5-3B $+\mu$Qwen2.5-32B-truncated, where $\mu$ is selected from $\{0.01,0.02,0.03,0.04,0.05\}$.

Although truncation is lossy, injecting the truncated 32B model with a small coefficient still provides a useful performance gain. Tables~\ref{tab:intersection-qwen25-3b}--\ref{tab:intersection-qwen3-4b} show that Qwen2.5-3B improves from 0.5459 to 0.5716, while Qwen2.5-14B improves from 0.7045 to 0.7161. The gains in Tables~\ref{tab:intersection-qwen25-14b}--\ref{tab:intersection-qwen3-4b} are milder because their dimensional gaps are smaller than the 3B--32B gap in Table~\ref{tab:intersection-qwen25-3b}. This suggests that intersection-style merging is most useful when the source models have a large capability gap, provided that the truncated branch remains a small perturbation.

\section{Analysis}

\subsection{Analysis of Gain Patterns}
As discussed in Section~5, gains concentrate when source models have partially distinct competence profiles. Figure~\ref{fig:radar-knowledge-structure} shows that intersecting but non-nested models leave transferable complementary regions, while the intersection-style results show that useful structure can survive truncation and be absorbed as a small perturbation (Table~\ref{tab:intersection-qwen25-3b}). The effect is more pronounced when the two models differ more substantially in scale.

\subsection{Ratio Sensitivity and Effective Region}
Direct averaging is effective mainly near an anchor model. Appendix~\ref{app:gsm8k} and Tables~\ref{tab:gsm8k-results}--\ref{tab:gsm8k-truncate-14b-results} show two protocol-specific phase patterns. For union-style merging:
\begin{equation}
\lambda \in (0,0.1) \cup (0.9,1) \Rightarrow \text{stable transfer},
\end{equation}
\begin{equation}
\lambda \in (0.1,0.9) \Rightarrow \text{severe performance collapse}.
\end{equation}
For intersection-style merging, where $\mu$ weights the truncated larger checkpoint:
\begin{equation}
\mu \in (0,0.1) \Rightarrow \text{stable transfer},
\end{equation}
\begin{equation}
\mu \in (0.1,1) \Rightarrow \text{severe performance collapse}.
\end{equation}
Thus, union-style merging has two endpoint neighborhoods, whereas intersection-style merging is one-sided because truncation is lossy. This ratio sensitivity is consistent with weight-space connectivity and alignment effects~\citep{garipov2018loss,frankle2020linear}.

\subsection{Seesaw Effect}
Heterogeneous merging is not monotonic. For Qwen2.5-14B and Qwen2.5-32B, the aggregate best merged configuration reaches 0.7525, above both sources (0.7044 and 0.7424; Table~\ref{tab:union-qwen25}), but the task-level summary in Appendix~\ref{app:task-level-results} shows that the worst average falls to 0.7036 with task-level drops (Table~\ref{tab:seesaw-qwen25}). Aggregate gains can therefore hide capability-specific regressions; Appendix~\ref{app:task-level-results} provides additional details in Tables~\ref{tab:seesaw-qwen3-4b-8b}--\ref{tab:seesaw-qwen3-4b-small-ratio}.

\subsection{Perturbation Controls and Baseline Comparisons}
We further test whether the gains can be explained by generic perturbation rather than cross-checkpoint transfer. Appendix~\ref{app:ablation-baselines} reports matched perturbation controls on the Qwen2.5-3B intersection setting. Target-only scaling and matched noise sometimes improve individual tasks, but they do not match the merged checkpoint on the reported tasks. For example, on RTE, the baseline, target-only scaling, matched noise, and merge scores are 0.6426, 0.6534, 0.6570, and 0.7256, respectively. On GSM8K, matched noise reaches 0.6808, while the merged checkpoint reaches 0.7028. This indicates that the improvement is not fully explained by adding a non-checkpoint perturbation of comparable scale.

We also compare weighted averaging after truncation with several standard merging rules in the same small-model parameter space. The comparison is not intended to show that weighted averaging dominates all possible merging algorithms. Instead, it tests whether the simplest zero-training, zero-alignment rule remains competitive after dimensional adaptation. On the reported average score, truncated-0.02 weighted averaging obtains 0.5551, close to TIES at 0.5587 and above SLERP, the 3B baseline, and DARE (Appendix~\ref{app:ablation-baselines}). Thus, dimensional adaptation should be viewed as a compatibility operator that can support simple averaging and may also support stronger or complementary merging rules.

\section{Conclusion}

We show that deterministic dimensional adaptation enables direct weighted averaging across heterogeneous LLM scales. Across Qwen-family model pairs, small-ratio union and intersection merging can preserve source behavior and transfer complementary capabilities without training, adapters, routing, or semantic alignment. These results make simple weighted fusion a competitive zero-training baseline for probing heterogeneous LLM merging. At the same time, ratio sensitivity, perturbation controls, and seesaw effects show that compatibility is local and depends on the source pair, merging direction, and evaluation task. Task-level validation therefore remains essential before treating a merged checkpoint as uniformly improved.

\section*{Limitations}

This study focuses on Qwen-family checkpoints, offline benchmarks, and representative ratios; broader evaluation of long-context, multilingual, tool-use, efficiency, calibration, robustness, and safety settings remains future work.

\section*{Ethical Considerations}

Merged models may inherit unsafe or biased behavior from source checkpoints, so deployment requires safety checks.


\onecolumn
\appendix
\clearpage
\Needspace{8\baselineskip}
\section{Additional Task-Level Results}
\label{app:task-level-results}

\begin{table}[!htbp]
\centering
\footnotesize
\setlength{\tabcolsep}{2.5pt}
\renewcommand{\arraystretch}{0.98}
\begin{tabularx}{\textwidth}{@{}>{\raggedright\arraybackslash}p{0.15\textwidth}|*{2}{>{\centering\arraybackslash}X}|*{4}{>{\centering\arraybackslash}X}|*{2}{>{\raggedleft\arraybackslash}X}@{}}
\toprule
\textbf{Dataset} & \textbf{14B} & \textbf{32B} & \textbf{$\lambda=0.98$} & \textbf{$\lambda=0.96$} & \textbf{$\lambda=0.04$} & \textbf{$\lambda=0.02$} & \textbf{BEST} & \textbf{WORST} \\
\midrule
\texttt{gsm8k} & 0.8779 & 0.9014 & \textbf{0.9045} & 0.8992 & 0.8787 & 0.8832 & \textbf{0.9045} & 0.8787 \\
\texttt{MATH-500} & 0.5460 & 0.6180 & 0.6140 & 0.5860 & 0.5380 & 0.5560 & 0.6140 & 0.5380 \\
\texttt{humanevalplus} & 0.5604 & 0.6165 & \textbf{0.6482} & \textbf{0.6494} & 0.5628 & 0.5591 & \bestcorner{0.6494}{0.03} & 0.5591 \\
\texttt{mbppplus} & 0.6738 & 0.7405 & \textbf{0.7413} & 0.7392 & 0.6791 & 0.6833 & \textbf{0.7413} & 0.6791 \\
\texttt{BoolQ} & 0.8532 & 0.8740 & 0.8719 & 0.8682 & 0.8554 & 0.8590 & 0.8719 & 0.8554 \\
\texttt{COPA} & 0.9000 & 0.8700 & 0.8900 & 0.8900 & 0.9000 & \textbf{0.9100} & \bestcorner{0.9100}{0.01} & 0.8900 \\
\texttt{MNLI} & 0.6689 & 0.7217 & \textbf{0.7257} & \textbf{0.7300} & 0.6869 & 0.6823 & \textbf{0.7300} & 0.6823 \\
\texttt{PIQA} & 0.8248 & 0.8243 & 0.8232 & 0.8243 & \textbf{0.8254} & 0.8226 & \textbf{0.8254} & 0.8226 \\
\texttt{QNLI} & 0.6771 & 0.8596 & 0.8587 & 0.8594 & 0.7428 & 0.7040 & 0.8594 & 0.7040 \\
\texttt{RTE} & 0.7906 & 0.8159 & 0.8123 & 0.8123 & 0.7870 & 0.7906 & 0.8123 & 0.7870 \\
\texttt{Winogrande} & 0.7569 & 0.7569 & 0.7569 & \textbf{0.7687} & \textbf{0.7774} & \textbf{0.7609} & \bestcorner{0.7774}{0.02} & 0.7569 \\
\texttt{ARC} & 0.5241 & 0.5320 & \textbf{0.5328} & \textbf{0.5341} & 0.5099 & 0.5144 & \textbf{0.5341} & 0.5099 \\
\texttt{HellaSwag} & 0.6911 & 0.7175 & 0.7156 & 0.7116 & 0.6844 & 0.6893 & 0.7156 & 0.6844 \\
\texttt{MMLU} & 0.7094 & 0.7620 & \textbf{0.7630} & \textbf{0.7623} & 0.7073 & 0.7091 & \textbf{0.7630} & 0.7073 \\
\texttt{ifeval} & 0.4424 & 0.4832 & \textbf{0.4880} & \textbf{0.5072} & 0.4317 & 0.4412 & \bestcorner{0.5072}{0.02} & 0.4317 \\
\texttt{BBH} & 0.7731 & 0.7852 & \textbf{0.8231} & \textbf{0.8241} & 0.7815 & 0.7704 & \bestcorner{0.8241}{0.04} & 0.7704 \\
\midrule
\textbf{Avg.} & 0.7044 & 0.7424 & \textbf{0.7481} & \textbf{0.7479} & 0.7093 & 0.7085 & \textbf{0.7525} & 0.7036 \\
\bottomrule
\end{tabularx}
\caption{Task-level best/worst outcomes for Qwen2.5-14B and Qwen2.5-32B under union-style merging.}
\label{tab:seesaw-qwen25}
\end{table}

\begin{table}[!htbp]
\centering
\footnotesize
\setlength{\tabcolsep}{2.5pt}
\renewcommand{\arraystretch}{0.98}
\begin{tabularx}{\textwidth}{@{}>{\raggedright\arraybackslash}p{0.15\textwidth}|*{2}{>{\centering\arraybackslash}X}|*{4}{>{\centering\arraybackslash}X}|*{2}{>{\raggedleft\arraybackslash}X}@{}}
\toprule
\textbf{Dataset} & \textbf{4B} & \textbf{8B} & \textbf{$\lambda=0.98$} & \textbf{$\lambda=0.96$} & \textbf{$\lambda=0.04$} & \textbf{$\lambda=0.02$} & \textbf{BEST} & \textbf{WORST} \\
\midrule
\texttt{gsm8k} & 0.8931 & 0.8764 & \textbf{0.9007} & 0.8749 & 0.8605 & 0.8923 & \textbf{0.9007} & 0.8605 \\
\texttt{MATH-500} & 0.5100 & 0.5300 & \textbf{0.5480} & \textbf{0.5460} & 0.5180 & 0.4680 & \bestcorner{0.5480}{0.02} & 0.4680 \\
\texttt{humanevalplus} & 0.5854 & 0.6091 & \textbf{0.6768} & \textbf{0.6646} & 0.5061 & 0.5915 & \bestcorner{0.6768}{0.07} & 0.5061 \\
\texttt{mbppplus} & 0.4550 & 0.6328 & 0.6296 & \textbf{0.6402} & 0.4497 & 0.4444 & \textbf{0.6402} & 0.4444 \\
\texttt{BoolQ} & 0.7477 & 0.8673 & \textbf{0.8682} & \textbf{0.8709} & 0.7569 & 0.7538 & \textbf{0.8709} & 0.7538 \\
\texttt{COPA} & 0.8200 & 0.8500 & \textbf{0.8700} & \textbf{0.8600} & 0.7900 & 0.8000 & \bestcorner{0.8700}{0.02} & 0.7900 \\
\texttt{MNLI} & 0.5849 & 0.6220 & \textbf{0.6298} & \textbf{0.6387} & 0.5849 & 0.5930 & \bestcorner{0.6387}{0.02} & 0.5849 \\
\texttt{PIQA} & 0.7546 & 0.7753 & \textbf{0.7775} & 0.7753 & 0.7481 & 0.7514 & \textbf{0.7775} & 0.7481 \\
\texttt{QNLI} & 0.5266 & 0.7824 & \textbf{0.7899} & \textbf{0.7842} & 0.5601 & 0.5435 & \textbf{0.7899} & 0.5435 \\
\texttt{RTE} & 0.8123 & 0.7762 & 0.7762 & 0.7834 & 0.8123 & 0.7978 & 0.8123 & 0.7762 \\
\texttt{Winogrande} & 0.6590 & 0.6765 & \textbf{0.6827} & \textbf{0.6851} & 0.6551 & 0.6567 & \textbf{0.6851} & 0.6551 \\
\texttt{ARC} & 0.3887 & 0.4640 & 0.4610 & 0.4601 & 0.3849 & 0.3868 & 0.4610 & 0.3849 \\
\texttt{HellaSwag} & 0.5152 & 0.6125 & 0.6094 & 0.6050 & 0.5039 & 0.5086 & 0.6094 & 0.5039 \\
\texttt{MMLU} & 0.5854 & 0.6695 & \textbf{0.6722} & \textbf{0.6716} & 0.5921 & 0.5918 & \textbf{0.6722} & 0.5918 \\
\texttt{ifeval} & 0.4652 & 0.4233 & 0.4077 & 0.3777 & \textbf{0.4880} & \textbf{0.4868} & \bestcorner{0.4880}{0.02} & 0.3777 \\
\texttt{BBH} & 0.7380 & 0.7889 & 0.7833 & 0.7852 & 0.7537 & 0.7509 & 0.7852 & 0.7509 \\
\midrule
\textbf{Avg.} & 0.6276 & 0.6848 & \textbf{0.6927} & \textbf{0.6889} & 0.6228 & 0.6261 & \bestcorner{0.7016}{0.02} & 0.6087 \\
\bottomrule
\end{tabularx}
\caption{Task-level best/worst outcomes for Qwen3-4B-Thinking-2507 and Qwen3-8B under union-style merging.}
\label{tab:seesaw-qwen3-4b-8b}
\end{table}

\begin{table}[!htbp]
\centering
\footnotesize
\setlength{\tabcolsep}{2.5pt}
\renewcommand{\arraystretch}{0.98}
\begin{tabularx}{\textwidth}{@{}>{\raggedright\arraybackslash}p{0.15\textwidth}|*{2}{>{\centering\arraybackslash}X}|*{4}{>{\centering\arraybackslash}X}|*{2}{>{\raggedleft\arraybackslash}X}@{}}
\toprule
\textbf{Dataset} & \textbf{14B} & \textbf{32B} & \textbf{$\lambda=0.98$} & \textbf{$\lambda=0.96$} & \textbf{$\lambda=0.04$} & \textbf{$\lambda=0.02$} & \textbf{BEST} & \textbf{WORST} \\
\midrule
\texttt{gsm8k} & 0.9227 & 0.9113 & 0.9181 & 0.8848 & 0.9121 & 0.9181 & 0.9181 & 0.8848 \\
\texttt{MATH-500} & 0.8160 & 0.7740 & 0.7740 & 0.7580 & 0.7920 & \textbf{0.8180} & \textbf{0.8180} & 0.7580 \\
\texttt{humanevalplus} & 0.6707 & 0.6890 & \textbf{0.7378} & 0.6890 & 0.6646 & 0.6585 & \bestcorner{0.7378}{0.05} & 0.6585 \\
\texttt{mbppplus} & 0.6878 & 0.6402 & 0.6587 & \textbf{0.6984} & \textbf{0.7169} & \textbf{0.7116} & \bestcorner{0.7169}{0.03} & 0.6587 \\
\texttt{BoolQ} & 0.8924 & 0.8633 & 0.8752 & 0.8789 & 0.8887 & 0.8896 & 0.8896 & 0.8752 \\
\texttt{COPA} & 0.9000 & 0.9400 & 0.9400 & \textbf{0.9600} & 0.8800 & 0.8900 & \bestcorner{0.9600}{0.02} & 0.8800 \\
\texttt{MNLI} & 0.6706 & 0.6989 & \textbf{0.7104} & 0.6951 & 0.6918 & 0.6847 & \bestcorner{0.7104}{0.01} & \textbf{0.6847} \\
\texttt{PIQA} & 0.7976 & 0.8150 & \textbf{0.8188} & 0.8025 & 0.8047 & 0.8009 & \textbf{0.8188} & \textbf{0.8009} \\
\texttt{QNLI} & 0.8380 & 0.8104 & 0.8365 & \textbf{0.8515} & \textbf{0.8543} & \textbf{0.8468} & \bestcorner{0.8543}{0.02} & 0.8365 \\
\texttt{RTE} & 0.7762 & 0.7545 & 0.7581 & 0.7653 & \textbf{0.7906} & \textbf{0.7870} & \bestcorner{0.7906}{0.01} & \textbf{0.7581} \\
\texttt{Winogrande} & 0.7277 & 0.7301 & 0.7261 & 0.7190 & 0.7222 & 0.7238 & 0.7261 & 0.7190 \\
\texttt{ARC} & 0.5093 & 0.5341 & \textbf{0.5352} & 0.5245 & 0.5178 & 0.5160 & \textbf{0.5352} & 0.5160 \\
\texttt{HellaSwag} & 0.6619 & 0.7065 & 0.7061 & 0.6782 & 0.6678 & 0.6655 & 0.7061 & 0.6655 \\
\texttt{MMLU} & 0.7135 & 0.7621 & 0.7604 & 0.7248 & 0.7099 & 0.7129 & 0.7604 & 0.7099 \\
\texttt{ifeval} & 0.4484 & 0.4652 & \textbf{0.5360} & \textbf{0.5635} & 0.4580 & \textbf{0.4688} & \bestcorner{0.5635}{0.10} & \textbf{0.4580} \\
\texttt{BBH} & 0.8500 & 0.8639 & \textbf{0.8843} & 0.8593 & 0.8583 & 0.8620 & \bestcorner{0.8843}{0.02} & 0.8583 \\
\midrule
\textbf{Avg.} & 0.7427 & 0.7474 & \textbf{0.7610} & \textbf{0.7533} & 0.7456 & 0.7471 & \bestcorner{0.7744}{0.03} & 0.7326 \\
\bottomrule
\end{tabularx}
\caption{Task-level best/worst outcomes for Qwen3-14B and Qwen3-32B under union-style merging.}
\label{tab:seesaw-qwen3-14b-32b}
\end{table}

\begin{table}[!htbp]
\centering
\footnotesize
\setlength{\tabcolsep}{2.7pt}
\renewcommand{\arraystretch}{1.05}
\begin{tabularx}{\textwidth}{@{}>{\raggedright\arraybackslash}p{0.15\textwidth}|>{\centering\arraybackslash}X|*{5}{>{\centering\arraybackslash}X}|*{2}{>{\raggedleft\arraybackslash}X}@{}}
\toprule
\textbf{Dataset} & \textbf{3B} & \textbf{$\mu=0.01$} & \textbf{$\mu=0.02$} & \textbf{$\mu=0.03$} & \textbf{$\mu=0.04$} & \textbf{$\mu=0.05$} & \textbf{BEST} & \textbf{WORST} \\
\midrule
\texttt{gsm8k} & 0.6566 & \textbf{0.6839} & \textbf{0.7028} & \textbf{0.7142} & \textbf{0.7210} & \textbf{0.7043} & \bestcorner{0.7210}{0.06} & \textbf{0.6839} \\
\texttt{MATH-500} & 0.3840 & \textbf{0.4060} & 0.3840 & 0.3560 & 0.3660 & 0.3560 & \bestcorner{0.4060}{0.02} & 0.3560 \\
\texttt{humanevalplus} & 0.4390 & 0.4329 & 0.3671 & 0.3232 & 0.3354 & 0.2927 & 0.4329 & 0.2927 \\
\texttt{mbppplus} & 0.5899 & 0.5873 & \textbf{0.6058} & 0.5847 & 0.5794 & \textbf{0.5926} & \bestcorner{0.6058}{0.02} & 0.5794 \\
\texttt{BoolQ} & 0.7080 & \textbf{0.7089} & 0.7040 & \textbf{0.7116} & \textbf{0.7254} & \textbf{0.7272} & \bestcorner{0.7272}{0.02} & 0.7040 \\
\texttt{COPA} & 0.7800 & \textbf{0.7900} & \textbf{0.7900} & \textbf{0.7900} & \textbf{0.8000} & \textbf{0.8000} & \bestcorner{0.8000}{0.02} & \textbf{0.7900} \\
\texttt{MNLI} & 0.4598 & \textbf{0.4817} & \textbf{0.4919} & \textbf{0.5037} & \textbf{0.5203} & \textbf{0.5334} & \bestcorner{0.5334}{0.07} & \textbf{0.4817} \\
\texttt{PIQA} & 0.7579 & 0.7563 & \textbf{0.7584} & 0.7481 & 0.7459 & 0.7405 & \textbf{0.7584} & 0.7405 \\
\texttt{QNLI} & 0.5286 & \textbf{0.5345} & 0.5279 & \textbf{0.5378} & \textbf{0.5451} & \textbf{0.5526} & \bestcorner{0.5526}{0.02} & 0.5279 \\
\texttt{RTE} & 0.6426 & \textbf{0.6931} & \textbf{0.7256} & \textbf{0.7581} & \textbf{0.7726} & \textbf{0.7437} & \bestcorner{0.7726}{0.13} & \textbf{0.6931} \\
\texttt{Winogrande} & 0.6338 & \textbf{0.6401} & 0.6322 & \textbf{0.6377} & 0.6212 & 0.6235 & \bestcorner{0.6401}{0.01} & 0.6212 \\
\texttt{ARC} & 0.3442 & 0.3430 & 0.3421 & 0.3372 & 0.3361 & 0.3357 & 0.3430 & 0.3357 \\
\texttt{HellaSwag} & 0.5094 & 0.5087 & 0.5068 & 0.5027 & 0.5003 & 0.4949 & 0.5087 & 0.4949 \\
\texttt{MMLU} & 0.4496 & \textbf{0.4504} & \textbf{0.4514} & 0.4407 & 0.4323 & 0.4241 & \textbf{0.4514} & 0.4241 \\
\texttt{ifeval} & 0.4041 & 0.3957 & \textbf{0.4113} & \textbf{0.4065} & 0.3861 & 0.3825 & \bestcorner{0.4113}{0.01} & 0.3825 \\
\texttt{BBH} & 0.4463 & \textbf{0.4731} & \textbf{0.4806} & \textbf{0.4694} & \textbf{0.4491} & 0.4306 & \bestcorner{0.4806}{0.03} & 0.4306 \\
\midrule
\textbf{Avg.} & 0.5459 & \textbf{0.5554} & \textbf{0.5551} & \textbf{0.5514} & \textbf{0.5523} & 0.5459 & \bestcorner{0.5716}{0.03} & 0.5336 \\
\bottomrule
\end{tabularx}
\caption{Task-level best and worst merging outcomes for Qwen2.5-3B under small-ratio intersection-style merging.}
\label{tab:seesaw-qwen25-3b-small-ratio}
\end{table}

\begin{table}[!htbp]
\centering
\footnotesize
\setlength{\tabcolsep}{2.7pt}
\renewcommand{\arraystretch}{1.05}
\begin{tabularx}{\textwidth}{@{}>{\raggedright\arraybackslash}p{0.15\textwidth}|>{\centering\arraybackslash}X|*{5}{>{\centering\arraybackslash}X}|*{2}{>{\raggedleft\arraybackslash}X}@{}}
\toprule
\textbf{Dataset} & \textbf{14B} & \textbf{$\mu=0.01$} & \textbf{$\mu=0.02$} & \textbf{$\mu=0.03$} & \textbf{$\mu=0.04$} & \textbf{$\mu=0.05$} & \textbf{BEST} & \textbf{WORST} \\
\midrule
\texttt{gsm8k} & 0.8802 & 0.8772 & \textbf{0.8848} & \textbf{0.8817} & \textbf{0.8810} & 0.8719 & \textbf{0.8848} & 0.8719 \\
\texttt{MATH-500} & 0.5460 & \textbf{0.5640} & \textbf{0.5520} & \textbf{0.5520} & 0.5460 & 0.5400 & \bestcorner{0.5640}{0.02} & 0.5400 \\
\texttt{humanevalplus} & 0.5604 & \textbf{0.5616} & 0.5579 & 0.5561 & 0.5494 & 0.5378 & \textbf{0.5616} & 0.5378 \\
\texttt{mbppplus} & 0.6738 & \textbf{0.6775} & \textbf{0.6767} & 0.6640 & 0.6624 & 0.6542 & \textbf{0.6775} & 0.6542 \\
\texttt{BoolQ} & 0.8532 & \textbf{0.8584} & \textbf{0.8612} & \textbf{0.8581} & \textbf{0.8560} & 0.8517 & \textbf{0.8612} & 0.8517 \\
\texttt{COPA} & 0.9000 & 0.9000 & \textbf{0.9100} & \textbf{0.9100} & \textbf{0.9100} & 0.9000 & \bestcorner{0.9100}{0.01} & 0.9000 \\
\texttt{MNLI} & 0.6689 & \textbf{0.6767} & \textbf{0.6801} & \textbf{0.6836} & \textbf{0.6864} & \textbf{0.6880} & \bestcorner{0.6880}{0.02} & \textbf{0.6767} \\
\texttt{PIQA} & 0.8248 & 0.8232 & \textbf{0.8254} & 0.8232 & 0.8237 & 0.8177 & \textbf{0.8254} & 0.8177 \\
\texttt{QNLI} & 0.6771 & \textbf{0.6943} & \textbf{0.7031} & \textbf{0.7262} & \textbf{0.7386} & \textbf{0.7657} & \bestcorner{0.7657}{0.09} & \textbf{0.6943} \\
\texttt{RTE} & 0.7906 & \textbf{0.7978} & \textbf{0.7978} & 0.7906 & 0.7906 & 0.7906 & \textbf{0.7978} & 0.7906 \\
\texttt{Winogrande} & 0.7569 & \textbf{0.7664} & \textbf{0.7601} & \textbf{0.7711} & \textbf{0.7774} & \textbf{0.7727} & \bestcorner{0.7774}{0.02} & \textbf{0.7601} \\
\texttt{ARC} & 0.5241 & 0.5200 & 0.5144 & 0.5123 & 0.5106 & 0.5050 & 0.5200 & 0.5050 \\
\texttt{HellaSwag} & 0.6911 & 0.6903 & 0.6895 & 0.6859 & 0.6845 & 0.6805 & 0.6903 & 0.6805 \\
\texttt{MMLU} & 0.7094 & 0.7087 & 0.7088 & 0.7079 & 0.7073 & 0.7047 & 0.7088 & 0.7047 \\
\texttt{ifeval} & 0.4424 & 0.4388 & 0.4353 & 0.4269 & 0.4281 & 0.4185 & 0.4388 & 0.4185 \\
\texttt{BBH} & 0.7731 & 0.7694 & 0.7694 & \textbf{0.7759} & \textbf{0.7852} & \textbf{0.7870} & \bestcorner{0.7870}{0.01} & 0.7694 \\
\midrule
\textbf{Avg.} & 0.7045 & \textbf{0.7078} & \textbf{0.7079} & \textbf{0.7078} & \textbf{0.7086} & \textbf{0.7054} & \bestcorner{0.7161}{0.01} & 0.6983 \\
\bottomrule
\end{tabularx}
\caption{Task-level best and worst merging outcomes for Qwen2.5-14B under small-ratio intersection-style merging.}
\label{tab:seesaw-qwen25-14b-small-ratio}
\end{table}

\begin{table}[!htbp]
\centering
\footnotesize
\setlength{\tabcolsep}{2.7pt}
\renewcommand{\arraystretch}{1.05}
\begin{tabularx}{\textwidth}{@{}>{\raggedright\arraybackslash}p{0.15\textwidth}|>{\centering\arraybackslash}X|*{5}{>{\centering\arraybackslash}X}|*{2}{>{\raggedleft\arraybackslash}X}@{}}
\toprule
\textbf{Dataset} & \textbf{14B} & \textbf{$\mu=0.01$} & \textbf{$\mu=0.02$} & \textbf{$\mu=0.03$} & \textbf{$\mu=0.04$} & \textbf{$\mu=0.05$} & \textbf{BEST} & \textbf{WORST} \\
\midrule
\texttt{gsm8k} & 0.9340 & \textbf{0.9363} & \textbf{0.9378} & \textbf{0.9386} & \textbf{0.9348} & 0.9242 & \textbf{0.9386} & 0.9242 \\
\texttt{MATH-500} & 0.8160 & 0.8160 & \textbf{0.8220} & \textbf{0.8280} & 0.8060 & 0.8100 & \bestcorner{0.8280}{0.01} & 0.8060 \\
\texttt{humanevalplus} & 0.7500 & \textbf{0.7622} & 0.7500 & \textbf{0.7561} & \textbf{0.7561} & \textbf{0.7683} & \bestcorner{0.7683}{0.02} & 0.7500 \\
\texttt{mbppplus} & 0.7381 & 0.7354 & \textbf{0.7434} & 0.7381 & \textbf{0.7407} & 0.7354 & \textbf{0.7434} & 0.7354 \\
\texttt{BoolQ} & 0.8924 & \textbf{0.8933} & 0.8920 & 0.8902 & 0.8902 & 0.8881 & \textbf{0.8933} & 0.8881 \\
\texttt{COPA} & 0.9000 & 0.9000 & 0.9000 & 0.9000 & 0.8900 & 0.8800 & 0.9000 & 0.8800 \\
\texttt{MNLI} & 0.6706 & 0.6699 & 0.6653 & 0.6633 & 0.6569 & 0.6533 & 0.6699 & 0.6533 \\
\texttt{PIQA} & 0.7976 & \textbf{0.8009} & \textbf{0.7982} & \textbf{0.8025} & \textbf{0.7998} & \textbf{0.7982} & \textbf{0.8025} & \textbf{0.7982} \\
\texttt{QNLI} & 0.8380 & 0.8376 & \textbf{0.8402} & \textbf{0.8420} & \textbf{0.8431} & \textbf{0.8446} & \textbf{0.8446} & 0.8376 \\
\texttt{RTE} & 0.7762 & \textbf{0.7834} & \textbf{0.7834} & \textbf{0.7942} & \textbf{0.7942} & \textbf{0.7978} & \bestcorner{0.7978}{0.02} & \textbf{0.7834} \\
\texttt{Winogrande} & 0.7277 & \textbf{0.7316} & \textbf{0.7332} & \textbf{0.7340} & 0.7269 & \textbf{0.7340} & \textbf{0.7340} & 0.7269 \\
\texttt{ARC} & 0.5093 & \textbf{0.5095} & \textbf{0.5121} & \textbf{0.5127} & \textbf{0.5112} & \textbf{0.5119} & \textbf{0.5127} & \textbf{0.5095} \\
\texttt{HellaSwag} & 0.6619 & \textbf{0.6642} & \textbf{0.6657} & \textbf{0.6661} & \textbf{0.6666} & \textbf{0.6683} & \textbf{0.6683} & \textbf{0.6642} \\
\texttt{MMLU} & 0.7135 & 0.7129 & 0.7117 & 0.7105 & 0.7095 & 0.7061 & 0.7129 & 0.7061 \\
\texttt{ifeval} & 0.4484 & \textbf{0.4664} & \textbf{0.4700} & \textbf{0.4616} & 0.4460 & \textbf{0.4712} & \bestcorner{0.4712}{0.02} & 0.4460 \\
\texttt{BBH} & 0.8500 & 0.8370 & 0.8426 & \textbf{0.8546} & \textbf{0.8602} & 0.8074 & \bestcorner{0.8602}{0.01} & 0.8074 \\
\midrule
\textbf{Avg.} & 0.7515 & \textbf{0.7535} & \textbf{0.7542} & \textbf{0.7558} & \textbf{0.7520} & 0.7499 & \bestcorner{0.7591}{0.01} & 0.7448 \\
\bottomrule
\end{tabularx}
\caption{Task-level best and worst merging outcomes for Qwen3-14B under small-ratio intersection-style merging.}
\label{tab:seesaw-qwen3-14b-small-ratio}
\end{table}

\begin{table}[!htbp]
\centering
\setlength{\abovecaptionskip}{2pt}
\setlength{\belowcaptionskip}{0pt}
\footnotesize
\setlength{\tabcolsep}{2.0pt}
\renewcommand{\arraystretch}{1.05}
\begin{tabularx}{\textwidth}{@{}>{\raggedright\arraybackslash}p{0.15\textwidth}|>{\centering\arraybackslash}X|*{5}{>{\centering\arraybackslash}X}|*{2}{>{\raggedleft\arraybackslash}X}@{}}
\toprule
\textbf{Dataset} & \textbf{4B} & \textbf{$\mu=0.01$} & \textbf{$\mu=0.02$} & \textbf{$\mu=0.03$} & \textbf{$\mu=0.04$} & \textbf{$\mu=0.05$} & \textbf{BEST} & \textbf{WORST} \\
\midrule
\texttt{gsm8k} & 0.8863 & 0.8772 & 0.8840 & 0.8817 & 0.8840 & 0.8749 & 0.8840 & 0.8749 \\
\texttt{MATH-500} & 0.5100 & \textbf{0.5240} & \textbf{0.5120} & \textbf{0.5380} & \textbf{0.5140} & \textbf{0.5200} & \bestcorner{0.5380}{0.03} & \textbf{0.5120} \\
\texttt{humanevalplus} & 0.5220 & 0.5055 & 0.4744 & 0.4476 & 0.4122 & 0.3933 & 0.5055 & 0.3933 \\
\texttt{mbppplus} & 0.5275 & 0.5193 & 0.5079 & 0.4942 & 0.4878 & 0.4452 & 0.5193 & 0.4452 \\
\texttt{BoolQ} & 0.7477 & \textbf{0.7636} & \textbf{0.7664} & \textbf{0.7758} & \textbf{0.7749} & \textbf{0.7798} & \bestcorner{0.7798}{0.03} & \textbf{0.7636} \\
\texttt{COPA} & 0.8200 & 0.8100 & 0.8000 & 0.7900 & 0.7800 & 0.7800 & 0.8100 & 0.7800 \\
\texttt{MNLI} & 0.5849 & \textbf{0.5879} & \textbf{0.5887} & \textbf{0.5881} & \textbf{0.5855} & 0.5828 & \textbf{0.5887} & 0.5828 \\
\texttt{PIQA} & 0.7546 & 0.7530 & 0.7524 & \textbf{0.7563} & \textbf{0.7557} & 0.7541 & \textbf{0.7563} & 0.7524 \\
\texttt{QNLI} & 0.5266 & \textbf{0.5347} & \textbf{0.5416} & \textbf{0.5599} & \textbf{0.5718} & \textbf{0.5863} & \bestcorner{0.5863}{0.06} & \textbf{0.5347} \\
\texttt{RTE} & 0.8123 & 0.8087 & 0.8014 & 0.8014 & 0.8087 & 0.8087 & 0.8087 & 0.8014 \\
\texttt{Winogrande} & 0.6590 & \textbf{0.6622} & \textbf{0.6646} & \textbf{0.6740} & \textbf{0.6748} & \textbf{0.6693} & \bestcorner{0.6748}{0.02} & \textbf{0.6622} \\
\texttt{ARC} & 0.3887 & 0.3872 & 0.3842 & 0.3840 & 0.3819 & 0.3791 & 0.3872 & 0.3791 \\
\texttt{HellaSwag} & 0.5152 & 0.5146 & 0.5136 & 0.5094 & 0.5081 & 0.5036 & 0.5146 & 0.5036 \\
\texttt{MMLU} & 0.5854 & \textbf{0.5923} & \textbf{0.5940} & \textbf{0.5919} & \textbf{0.5912} & \textbf{0.5908} & \textbf{0.5940} & \textbf{0.5908} \\
\texttt{ifeval} & 0.4652 & \textbf{0.4700} & 0.4592 & \textbf{0.4772} & \textbf{0.4700} & \textbf{0.4748} & \bestcorner{0.4772}{0.01} & 0.4592 \\
\texttt{BBH} & 0.7380 & \textbf{0.7574} & \textbf{0.7574} & \textbf{0.7574} & \textbf{0.7657} & \textbf{0.7519} & \bestcorner{0.7657}{0.03} & \textbf{0.7519} \\
\midrule
\textbf{Avg.} & 0.6277 & \textbf{0.6292} & 0.6251 & 0.6267 & 0.6229 & 0.6184 & \bestcorner{0.6369}{0.01} & 0.6117 \\
\bottomrule
\end{tabularx}
\caption{Task-level best and worst merging outcomes for Qwen3-4B-Thinking-2507 under small-ratio intersection-style merging.}
\label{tab:seesaw-qwen3-4b-small-ratio}
\end{table}

\Needspace{8\baselineskip}
\section{Dimensional Adaptation Details}
\label{app:method-details}

This appendix gives the implementation-level form of the two dimensional adaptation operators used in Section~3. We separate the operator that maps a smaller checkpoint into a larger architecture, denoted by \textsc{Expand}, from the operator that projects a larger checkpoint into a smaller architecture, denoted by \textsc{Truncate}. These operators correspond to union-style merging and intersection-style merging, respectively. They are deterministic tensor maps applied before interpolation, not learned alignment modules.

\subsection{Union-Style \textsc{Expand}}

Let the source checkpoint have hidden size $h_s$, intermediate size $m_s$, number of layers $L_s$, number of query heads $H_s$, number of key-value heads $K_s$, and head dimension $d_h$. Let the target architecture have the corresponding quantities $h_t,m_t,L_t,H_t,K_t$, with target dimensions no smaller than the source dimensions for the expanded axes. We use a slot map
\begin{equation}
\pi_{n\rightarrow m}(i)=\left\lfloor \frac{i m}{n}\right\rfloor,\qquad i=0,\ldots,n-1,
\end{equation}
to place source coordinates or heads into deterministic target slots. For a tensor $W\in\mathbb{R}^{a_s\times b_s}$ expanded along rows and columns, the generic zero-extension rule is
\begin{equation}
\big[\mathcal{E}(W)\big]_{\pi_{a_s\rightarrow a_t}(i),\,\pi_{b_s\rightarrow b_t}(j)}
= c\,W_{ij},\qquad
\big[\mathcal{E}(W)\big]_{uv}=0\ \text{otherwise},
\end{equation}
where $c$ is one for ordinary linear weights and may be set to $\sqrt{h_s/h_t}$ for copied hidden-width normalization coordinates in width-changing expansions. In practice, hidden coordinates are placed at a head-dimension block granularity when the architecture exposes a fixed head dimension. This preserves contiguous per-head subspaces rather than scattering individual coordinates.

Attention projections are expanded with grouped-query structure preserved. Query rows are mapped at whole-head granularity, while key and value rows are mapped at key-value-head granularity. The output projection applies the inverse placement on its input-head dimension and the hidden-coordinate placement on its output dimension. Thus the target architecture keeps its own attention scale $1/\sqrt{d_h}$ and head dimension. New attention rows or columns are zero at initialization, so they do not introduce untuned high-variance logits before interpolation.

The MLP expansion follows the semantic axes of the SwiGLU block. For gate and up projections, source rows are copied into the first or mapped $m_s$ intermediate slots and source hidden columns are copied into mapped hidden slots. For the down projection, source hidden rows are copied into mapped hidden slots and source intermediate columns are copied into the retained intermediate slots. All newly introduced MLP coordinates are zero-filled. Because a SwiGLU branch has the form
\begin{equation}
\operatorname{MLP}(x)=W_{\mathrm{down}}\left(\operatorname{SwiGLU}(W_{\mathrm{gate}}x,W_{\mathrm{up}}x)\right),
\end{equation}
zero-filling new gate, up, and down coordinates prevents newly inserted nonlinear branches from contributing before interpolation.

For additional layers, we use residual identity initialization. Consider a pre-normalization Transformer block with normalization operator $\mathcal{N}_{\ell}$ and attention/MLP branch $\mathcal{G}_{\ell}$:
\begin{equation}
h_{\ell+1}=h_{\ell}+\mathcal{G}_{\ell}\big(\mathcal{N}_{\ell}(h_{\ell});W_{\ell}\big).
\end{equation}
For a newly inserted layer, all attention and MLP branch weights are set to zero, while normalization scale parameters in the inserted layer are set to the neutral value one. This gives
\begin{equation}
\mathcal{G}_{\ell}\big(\mathcal{N}_{\ell}(h_{\ell});W_{\ell}^{\mathrm{new}}\big)=0,
\qquad h_{\ell+1}=h_{\ell},
\end{equation}
so $\Delta h_{\ell}=0$ at initialization. Normalization scales are not set to zero, because zero scales would collapse normalized feature variation. For LayerNorm~\citep{ba2016layer},
\begin{equation}
\mathcal{N}_{\ell}(h)=\gamma_{\ell}\odot
\frac{h-\bar{h}}{\sqrt{\operatorname{Var}(h)+\epsilon}}+\beta_{\ell},
\end{equation}
and the local Jacobian scales with $\gamma_{\ell}$. Setting $\gamma_{\ell}=\mathbf{1}$ preserves a well-conditioned normalized signal. The same intuition applies to RMSNorm~\citep{zhang2019root}.

\begin{table}[!t]
\centering
\small
\setlength{\tabcolsep}{4pt}
\renewcommand{\arraystretch}{1.12}
\begin{tabularx}{\textwidth}{@{}rX@{}}
\toprule
\multicolumn{2}{@{}l}{\textbf{Pseudocode 1: \textsc{Expand} for union-style merging}}\\
\midrule
\multicolumn{2}{@{}l}{\textbf{Input:} source checkpoint $\theta_s$, target architecture $\mathcal{A}_t$, source and target configs.}\\
\multicolumn{2}{@{}l}{\textbf{Output:} expanded checkpoint $\tilde{\theta}_s=\mathcal{E}(\theta_s)$ in the target parameter space.}\\
1 & Initialize a model with architecture $\mathcal{A}_t$ and obtain its target state dictionary.\\
2 & Build slot maps for hidden coordinates, query heads, key-value heads, and intermediate MLP dimensions.\\
3 & Copy embeddings and the language-model head into mapped hidden coordinates; set unmapped coordinates to zero.\\
4 & For each source layer $\ell<L_s$, copy normalization parameters into matched coordinates and initialize newly introduced normalization scales to neutral values when applicable.\\
5 & Expand $q$, $k$, $v$, and $o$ attention matrices by whole-head or key-value-head slots, preserving grouped-query attention structure.\\
6 & Expand MLP gate/up/down matrices along their semantic input and output dimensions; zero-fill newly introduced SwiGLU coordinates.\\
7 & For layers $L_s,\ldots,L_t-1$, set attention and MLP branch weights to zero and set normalization scales to one, producing residual identity blocks.\\
8 & Save the expanded checkpoint and use it in $\theta^{\cup}(\lambda)=(1-\lambda)\mathcal{E}(\theta_s)+\lambda\theta_t$.\\
\bottomrule
\end{tabularx}
\caption{\textsc{Expand} constructs a larger-architecture checkpoint by deterministic slot insertion, zero-filled new branches, and residual identity initialization for added layers.}
\label{alg:expand}
\end{table}

The stability claim for union-style merging is local. After expansion, interpolation can be written as
\begin{equation}
\theta^{\cup}(\lambda)=\mathcal{E}(\theta_s)+\lambda\big(\theta_t-\mathcal{E}(\theta_s)\big).
\end{equation}
For small $\lambda$, the expanded source checkpoint remains the anchor. For $\lambda$ close to one, the target checkpoint becomes the anchor. Near-balanced interpolation does not have this protection, and nonlinear gates can amplify destructive interference. This explains why the method can work in endpoint neighborhoods while failing in the middle of the ratio range.

\subsection{Intersection-Style \textsc{Truncate}}

For intersection-style merging, we instantiate the small-model architecture and project the large checkpoint into that state dictionary. Let $W_B\in\mathbb{R}^{s_1\times\cdots\times s_r}$ be a tensor from the larger checkpoint and let $W_A\in\mathbb{R}^{t_1\times\cdots\times t_r}$ be the corresponding tensor shape in the smaller architecture. The truncation operator copies the overlapping prefix region:
\begin{equation}
\big[\mathcal{T}(W_B)\big]_{i_1,\ldots,i_r}
=
\big[W_B\big]_{i_1,\ldots,i_r},
\qquad
0\le i_j<\min(s_j,t_j).
\end{equation}
Dimensions outside the target shape are discarded. Identical-shape tensors are copied directly. Layers whose indices exceed the small model depth are removed. For MLP matrices, this corresponds to retaining the target intermediate width: gate and up projections are sliced on their output dimension, while down projections are sliced on their input-intermediate dimension. For tied embeddings, the output head follows the target small-model tying convention.

\begin{table}[!t]
\centering
\small
\setlength{\tabcolsep}{4pt}
\renewcommand{\arraystretch}{1.12}
\begin{tabularx}{\textwidth}{@{}rX@{}}
\toprule
\multicolumn{2}{@{}l}{\textbf{Pseudocode 2: \textsc{Truncate} for intersection-style merging}}\\
\midrule
\multicolumn{2}{@{}l}{\textbf{Input:} large checkpoint $\theta_B$, small architecture $\mathcal{A}_A$, small tokenizer/config metadata.}\\
\multicolumn{2}{@{}l}{\textbf{Output:} projected checkpoint $\tilde{\theta}_B=\mathcal{T}(\theta_B)$ in the small parameter space.}\\
1 & Initialize a model with the small architecture $\mathcal{A}_A$ and read its target state-dictionary keys and shapes.\\
2 & For each target tensor key, find the tensor with the same key in the large checkpoint.\\
3 & If the source and target shapes are identical, copy the source tensor with the target dtype.\\
4 & If the ranks match but some dimensions are larger in the source tensor, copy the overlapping prefix slice into the target tensor.\\
5 & If a large-model layer index is beyond the target depth, discard that layer.\\
6 & If a target key is tied to the embedding matrix, reuse the target embedding tensor according to the small-model convention.\\
7 & Save the projected checkpoint and tokenizer metadata from the small architecture.\\
8 & Use the projection only as a perturbation in $\theta^{\cap}(\mu)=(1-\mu)\theta_A+\mu\mathcal{T}(\theta_B)$ with small $\mu$.\\
\bottomrule
\end{tabularx}
\caption{\textsc{Truncate} projects the large checkpoint into the small-model parameter space by same-key copying, prefix slicing of oversized tensors, and layer removal.}
\label{alg:truncate}
\end{table}

Truncation is intentionally lossy. We therefore do not require $\mathcal{T}(\theta_B)$ to be a strong standalone model. The merged checkpoint can be written as
\begin{equation}
\theta^{\cap}(\mu)=\theta_A+\mu\big(\mathcal{T}(\theta_B)-\theta_A\big),
\end{equation}
so the residual stream, normalization scales, and attention computations remain anchored by $\theta_A$ when $\mu$ is small. If the local input-output map around $\theta_A$ is approximated by a locally Lipschitz function $F$, then
\begin{equation}
\left\|F(\theta^{\cap}(\mu))-F(\theta_A)\right\|
\lesssim
\mu L \left\|\mathcal{T}(\theta_B)-\theta_A\right\|,
\end{equation}
up to higher-order nonlinear terms. This bound is only a local stability interpretation, not a guarantee. In particular, SwiGLU gates, residual-stream scale changes, and attention-logit entropy can make the middle-ratio regime unstable. Our experiments therefore treat attention entropy, task-level regressions, and collapse at $\mu\in(0.1,0.9)$ as diagnostics of the compatibility boundary rather than as contradictions of the small-ratio result.

The absence of permutation or semantic alignment is deliberate. \textsc{Expand} and \textsc{Truncate} assume architecture-compatible cross-scale models whose tensor names, module roles, and head structure are sufficiently comparable. The collapse regime marks where this assumption becomes too weak for direct weighted averaging.

\clearpage
\Needspace{8\baselineskip}
\section{Additional Diagnostics}
\label{app:ablation-baselines}
\label{app:gsm8k}

\subsection{Ablation Controls and Baseline Comparisons}

We add two diagnostic comparisons to separate cross-checkpoint transfer from simpler alternative explanations. Table~\ref{tab:perturbation-controls} compares intersection-style merging with target-only scaling and matched noise controls in the Qwen2.5-3B setting. Table~\ref{tab:baseline-comparison} compares weighted averaging after truncation with representative merging rules evaluated in the same small-model parameter space.

\begin{table}[H]
\centering
\small
\setlength{\tabcolsep}{5pt}
\renewcommand{\arraystretch}{1.12}
\begin{tabularx}{0.92\textwidth}{@{}lXrrrr@{}}
\toprule
\textbf{Setting} & \textbf{Task} & \textbf{Baseline} & \textbf{Target-only scaling} & \textbf{Matched noise} & \textbf{Merge} \\
\midrule
Intersection & \texttt{gsm8k} & 0.6566 & 0.6581 & 0.6808 & \textbf{0.7028} \\
Intersection & \texttt{mbppplus} & 0.5899 & 0.5847 & 0.5767 & \textbf{0.6058} \\
Intersection & \texttt{MNLI} & 0.4598 & 0.4586 & 0.4691 & \textbf{0.4919} \\
Intersection & \texttt{PIQA} & 0.7579 & 0.7568 & 0.7530 & \textbf{0.7584} \\
Intersection & \texttt{RTE} & 0.6426 & 0.6534 & 0.6570 & \textbf{0.7256} \\
Intersection & \texttt{MMLU} & 0.4496 & 0.4437 & 0.4499 & \textbf{0.4514} \\
\midrule
Union diagnostic & \texttt{IFEval} & 0.4652 & \multicolumn{2}{c}{0.4820 (2\% noise)} & \textbf{0.5360} at $\lambda=0.98$ \\
\bottomrule
\end{tabularx}
\caption{Perturbation controls for the Qwen2.5-3B intersection setting and a union diagnostic on IFEval. Target-only scaling and matched noise test whether gains can be explained by non-checkpoint perturbations of comparable scale.}
\label{tab:perturbation-controls}
\end{table}

\begin{table}[H]
\centering
\small
\setlength{\tabcolsep}{8pt}
\renewcommand{\arraystretch}{1.12}
\begin{tabular}{@{}lr@{}}
\toprule
\textbf{Method} & \textbf{Average score} \\
\midrule
TIES & \textbf{0.5587} \\
Truncated-0.02 weighted averaging & 0.5551 \\
SLERP & 0.5474 \\
3B baseline & 0.5459 \\
DARE & 0.5110 \\
\bottomrule
\end{tabular}
\caption{Baseline comparison in the small-model parameter space. Weighted averaging is not claimed to dominate all merging rules, but remains competitive as a zero-training, zero-alignment baseline after truncation.}
\label{tab:baseline-comparison}
\end{table}

\Needspace{8\baselineskip}
\subsection{GSM8K Ratio Sensitivity}

\begin{table}[!h]
\centering
\begin{minipage}[t]{0.48\textwidth}
\raggedright
\captionsetup{justification=raggedright,singlelinecheck=false,width=\linewidth}
\small
\setlength{\tabcolsep}{6pt}
\renewcommand{\arraystretch}{1.2}
\begin{tabular}{@{}l|cc|cc@{}}
\toprule
\textbf{Model/Ratio} & \multicolumn{2}{c|}{\textbf{Flexible}} & \multicolumn{2}{c}{\textbf{Strict}} \\
& \small Mean & \small Var & \small Mean & \small Var \\
\midrule
Qwen2.5-32B & 0.9014 & 0.0082 & 0.8021 & 0.0110 \\
Qwen2.5-14B & 0.8779 & 0.0090 & 0.7339 & 0.0122 \\
Expand & 0.8802 & 0.0089 & 0.7346 & 0.0122 \\
\midrule
$\lambda=0.1$ & 0.8059 & 0.0109 & 0.7210 & 0.0124 \\
$\lambda=0.2$ & 0.0781 & 0.0074 & 0.0197 & 0.0038 \\
$\lambda=0.3$ & 0.0144 & 0.0033 & 0.0000 & 0.0000 \\
$\lambda=0.4$ & 0.0045 & 0.0019 & 0.0000 & 0.0000 \\
$\lambda=0.5$ & 0.0121 & 0.0030 & 0.0000 & 0.0000 \\
$\lambda=0.6$ & 0.0227 & 0.0041 & 0.0023 & 0.0013 \\
$\lambda=0.7$ & 0.4496 & 0.0137 & 0.2919 & 0.0125 \\
$\lambda=0.8$ & 0.8597 & 0.0096 & 0.6285 & 0.0133 \\
$\lambda=0.9$ & 0.9045 & 0.0081 & 0.6785 & 0.0129 \\
\bottomrule
\end{tabular}
\caption{GSM8K ratio sensitivity for union-style merging.}
\label{tab:gsm8k-results}
\end{minipage}
\hfill
\begin{minipage}[t]{0.48\textwidth}
\raggedright
\captionsetup{justification=raggedright,singlelinecheck=false,width=\linewidth}
\small
\setlength{\tabcolsep}{6pt}
\renewcommand{\arraystretch}{1.2}
\begin{tabular}{@{}l|cc|cc@{}}
\toprule
\textbf{Model/Ratio} & \multicolumn{2}{c|}{\textbf{Flexible}} & \multicolumn{2}{c}{\textbf{Strict}} \\
& \small Mean & \small Var & \small Mean & \small Var \\
\midrule
Qwen2.5-32B & 0.9014 & 0.0082 & 0.8021 & 0.0110 \\
Qwen2.5-14B & 0.8779 & 0.0090 & 0.7339 & 0.0122 \\
Truncate & 0.0114 & 0.0029 & 0.0000 & 0.0000 \\
\midrule
$\mu=0.1$ & 0.8173 & 0.0106 & 0.7445 & 0.0120 \\
$\mu=0.2$ & 0.2176 & 0.0114 & 0.2146 & 0.0113 \\
$\mu=0.3$ & 0.0220 & 0.0040 & 0.0000 & 0.0000 \\
$\mu=0.4$ & 0.0159 & 0.0034 & 0.0000 & 0.0000 \\
$\mu=0.5$ & 0.0076 & 0.0024 & 0.0000 & 0.0000 \\
$\mu=0.6$ & 0.0000 & 0.0000 & 0.0000 & 0.0000 \\
$\mu=0.7$ & 0.0000 & 0.0000 & 0.0000 & 0.0000 \\
$\mu=0.8$ & 0.0053 & 0.0020 & 0.0000 & 0.0000 \\
$\mu=0.9$ & 0.0053 & 0.0020 & 0.0000 & 0.0000 \\
\bottomrule
\end{tabular}
\caption{GSM8K ratio sensitivity for intersection-style merging.}
\label{tab:gsm8k-truncate-14b-results}
\end{minipage}
\end{table}

\end{document}